\documentclass[journal]{IEEEtran}
\usepackage{blindtext}
\usepackage{array}
\usepackage{mdwmath}
\usepackage{mdwtab}
\usepackage{url}
\usepackage{color}
\usepackage[]{graphicx}
\usepackage{epsfig}
\usepackage{bbm}
\usepackage{enumerate}
\usepackage{amsfonts}
\usepackage{amsmath}
\usepackage{amsthm}
\usepackage{amssymb}
\usepackage{cite}
\usepackage{algorithm}
\usepackage{algpseudocode}
\usepackage{eqparbox}
\usepackage{comment}
\usepackage[tight,footnotesize]{subfigure}
\usepackage{float}
\usepackage{mathtools}
\usepackage{centernot}
\usepackage{gensymb}
\usepackage{caption}
\usepackage{lipsum}
\usepackage{multicol}

\DeclareMathOperator*{\argmin}{argmin}

\renewcommand{\algorithmiccomment}[1]{\hfill \eqparbox{COMMENT}{// #1}}

\usepackage{adjustbox}
\usepackage{lettrine}

\usepackage[section]{placeins}

\renewcommand{\algorithmiccomment}[1]{\hfill //~#1\normalsize}

\usepackage{doi}

\newtheorem{theorem}{Theorem}[section]

\newtheorem{lemma}[theorem]{Lemma}
\usepackage[utf8]{inputenc}
\usepackage[english]{babel}

\begin{document}

\title{Plug-and-Play Priors for Bright Field Electron Tomography and Sparse Interpolation}

\author{Suhas~Sreehari*,~\IEEEmembership{Student~Member,~IEEE,}
        S.~V.~Venkatakrishnan,~\IEEEmembership{Member,~IEEE,}
        Brendt Wohlberg,~\IEEEmembership{Senior Member,~IEEE,}
        Lawrence~F.~Drummy,
        Jeffrey~P.~Simmons,~\IEEEmembership{Member,~IEEE,}
        and~Charles~A.~Bouman,~\IEEEmembership{Fellow,~IEEE}
        
        \thanks{S. Sreehari and C. A. Bouman are with the School
of Electrical and Computer Engineering, Purdue University, West Lafayette,
IN, 47907 USA *E-mail: ssreehar@purdue.edu}
\thanks{S. V. Venkatakrishnan is with Lawrence Berkeley National Laboratory, Berkeley, CA.}
\thanks{B. Wohlberg is with Theoretical Division, Los Alamos National Laboratory, Los Alamos, NM.}
\thanks{L. F. Drummy and J. P Simmons are with Air Force Research Laboratory, Dayton, OH.}

\thanks{This work was supported by an AFOSR/MURI grant  \#FA9550-12-1-0458, 
by UES Inc. under the Broad Spectrum Engineered Materials contract, 
and by the Electronic Imaging component of the ICMD program 
of the Materials and Manufacturing Directorate of the Air Force Research Laboratory, Andrew Rosenberger, program manager.}
}


\maketitle

\begin{abstract}

Many material and biological samples in scientific imaging 
are characterized by non-local repeating structures. 
These are studied using scanning electron microscopy and electron tomography.
Sparse sampling of individual pixels in a 2D image acquisition geometry, or sparse sampling of projection images with large tilt increments in a tomography experiment, can enable high speed data acquisition and minimize sample damage caused by the electron beam.  

In this paper, we present an algorithm for electron tomographic reconstruction 
and sparse image interpolation that exploits the non-local redundancy in images. 
We adapt a framework, termed plug-and-play (P\&P) priors, 
to solve these imaging problems in a regularized inversion setting. 
The power of the P\&P approach is that it allows a wide array of modern denoising algorithms to be used as a ``prior model'' for tomography and image interpolation.
We also present sufficient mathematical conditions that ensure convergence of the P\&P approach, and we use these insights to design a new non-local means denoising algorithm. 
Finally, we demonstrate that the algorithm produces higher quality reconstructions on both simulated and real electron microscope data, along with improved convergence properties compared to other methods.

\end{abstract}

\begin{IEEEkeywords}
Plug-and-play, prior modeling, bright field electron tomography, sparse interpolation, non-local means, doubly-stochastic gradient non-local means, BM3D.
\end{IEEEkeywords}

\IEEEpeerreviewmaketitle

\section{Introduction}
\label{section:Introduction}

\lettrine{T}{ransmission} electron microscopes are widely used for characterization 
of material and biological samples at the nano-meter scale \cite{grunewald2003three, BarcETLifeScience, MidgleyBorkowski_2009_Review}. 
In many cases, these electron microscopy samples contain many repeating structures 
that are similar or identical to each other. 
High quality reconstruction of these samples from tomographic projections is 
possible by exploiting the redundancy caused by repeating structures.
As an important example, cryo-electron microscope (EM) tomography involves 
single particle reconstructions using several views 
of the same particle \cite{grunewald2003three}.
However, in the more general area of 3D transmission electron microscopy 
(TEM) tomography, no solution currently exists to 
fully exploit the redundancy in images constituted by many similar or identical particles.

Another important imaging problem is that raster scanning an electron beam across a 
large field of view is time consuming and can damage the sample.
For this reason, there is growing interest in reconstructing full resolution images 
from sparsely sampled pixels \cite{hyrumEI13, carinSparseInterp14}. 
The redundancy in material and biological samples suggests 
that it is possible to reconstruct such images with sufficient fidelity  
by acquiring only a few random samples in the image and using an advanced image 
reconstruction algorithm that exploits non-local redundancies.

Conventionally, model-based iterative reconstruction (MBIR) 
solves a single optimization problem that tightly couples the log likelihood term 
(based on the data) and the log of the prior probability 
\cite{DjafariJointMAP, BouSauGGMRF, Bo641Text, Sullivan:14, SeungseokMultiGrid, MumcuConjGrad, husarik2012radiation, venkatakrishnan2013TIP, venkatakrishnan2014TCI}. 
MBIR can, in principle, exploit redundancy in microscope images for tomographic reconstruction. 
This requires selection of the appropriate log prior probability, which is 
very challenging in practice.
Patch-based denoising algorithms such as non-local 
means (NLM) \cite{buades2005review, wong2008nonlocal, mairal2009non} 
and BM3D \cite{DabovBM3D07} have been very successful 
in exploiting non-local redundancy in images.
However, since NLM and BM3D are not explicitly formulated
as cost functions, it is unclear how to use them as prior models in the MBIR framework. Venkatakrishnan et al. \cite{venkatakrishnan2013school} 
developed a semi-empirical framework termed plug-and-play priors 
to incorporate such algorithms into general inverse problems, but 
limited results were presented and the convergence of the algorithm was not discussed.
Chen et al. \cite{chen2008nonlocal} proposed an MRF-style prior, but with non-local spatial dependencies, 
to perform Bayesian tomographic reconstruction.
The authors adopted a two-step optimization involving non-local weight update, 
followed by the image update. 
However, the cost function changes every iteration, so that there is no single fixed cost function that is minimized.
Chun et al. \cite{chun2013alternating} proposed non-local regularizers for emission tomography based on alternating direction method 
of multipliers (ADMM) \cite{glowinski1975approximation, gabay1976dual, eckstein1992douglas, boyd2011distributed}, 
using Fair potential \cite{fair1974robust} as the non-local regularizer, 
instead of non-local means. 
This model is restricted to convex potential functions, 
which in practice is a very strong constraint, and severely limits how expressive the model can be.
Yang et al. proposed a unifying energy minimization framework for non-local regularization \cite{yang2013nonlocal}, 
resulting in a model that captures the intrinsically non-convex behavior required for modeling distant particles with similar structure. 
However, it is not clear under what conditions their method converges.
Non-local regularizers using PDE-like evolutions and total variation are proposed to solve inverse problems \cite{gilboa2008nonlocal, peyre2008non}.

Image interpolation is also a widely 
researched problem \cite{inpaintReviewSPM14}. 
The approaches can be broadly classified into two 
categories - those based on local regularization 
and those on non-local regularization. 
In local approaches, the missing pixels are reconstructed from an 
immediate neighborhood 
surrounding the unknown values to encourage similarity 
between spatially neighboring pixels \cite{sapiroSiggraph00}. 
Spurred by the success of non-local means, 
there have been several efforts to solve the sparse interpolation problem using 
global patch based dictionary models
\cite{eladHarAna05, dong2013sparse, mairalJMLR10, carinSparseInterp14, sapiroGMMTIP15}.
Li et al. \cite{liTIP14} adapted a two stage approach
similar to \cite{DanEgi12} 
to the problem of sparse image reconstruction using the BM3D denoising algorithm.  
However, this approach is not immediately
applicable to denoising operators such as NLM and those 
formulated using a
nonparametric point estimation framework 
\cite{PeymanTour13, PeymanPLOW12, talebiGlideTIP14}.
The simplicity and success of NLM and 
BM3D has also led to the 
question of how these algorithms can be used to solve other inverse problems. 
In fact, Danielyan et al. \cite{DanEgi12} have adapted BM3D 
for image deblurring through the optimization 
of two cost functions balanced by the generalized Nash equilibrium.

In this paper, we present an algorithm for tomographic reconstruction 
and sparse image interpolation that exploits the non-local 
redundancies in microscope images. 
Our solution uses the plug-and-play (P\&P) framework 
\cite{venkatakrishnan2013school, sreehari2015advanced} 
which is based on the alternating direction method of multipliers 
(ADMM) \cite{gabay1976dual, eckstein1992douglas} and decouples the forward model  
and the prior terms in the optimization procedure. 
This results in an algorithm that involves repeated application of two steps: 
an inversion step only dependent on the forward model, and a denoising 
step only dependent on the image prior model.
The P\&P takes ADMM one step further by replacing the prior model optimization by a denoising operator.
However, while it is convenient to be able to use any denoising operator as a prior model,
this new framework also begs the question as to whether P\&P necessarily inherits  
the convergence properties of ADMM?
We answer this important question by presenting a theorem that outlines 
the sufficiency conditions to be satisfied by the denoising operator 
in order to guarantee convergence of the P\&P algorithm. 
We also present a proof for this convergence theorem partly based on the ideas presented by Moreau \cite{moreau1965proximite} and Williamson et al. \cite{vector_calculus_book}.
Using this result, we then modify NLM to satisfy these sufficiency conditions 
and call it doubly-stochastic gradient NLM (DSG-NLM).
We then apply DSG-NLM as a prior model 
to the tomographic reconstruction and sparse interpolation problems. 
This new DSG-NLM algorithm is based on symmetrizing the filter corresponding to the 
traditional NLM algorithm. 
Interestingly, Milanfar \cite{milanfar2013symmetrizing} has 
also discussed the benefit of symmetrizing the denoising operator, 
albeit in the context of improving the performance 
of heuristic denoising algorithms.

The plug-and-play electron tomography solution presented in this paper builds on the existing MBIR framework for bright field electron tomography \cite{venkatakrishnan2014TCI}, 
which models Bragg scatter and anomaly detection.
We demonstrate that our proposed algorithm produces high quality 
tomographic reconstructions and interpolation on 
both simulated and real electron microscope images. 
Additionally our method has improved convergence properties compared to using the standard NLM or the BM3D algorithm as a regularizer for the reconstruction.
Due to the generality of the plug-and-play technique, this work results 
in an MBIR framework that is compatible with any denoising algorithm as a prior model, 
and thus opens up a huge opportunity to adopt a wide variety of spatial constraints to solve a wide variety of inverse problems.

\section{Plug-and-play framework}
\label{section:Foundations}

Let $x \in \mathbb{R}^N$ be an unknown image with a prior distribution given by $p(x)$,
and let $y \in \mathbb{R}^M$ be the associated measurements of the image
with conditional distribution given by $p(y|x)$.
We will refer to $p(y|x)$ as the forward model for the measurement system.
Then the maximum \textit{a posteriori} (MAP) estimate of the image $x$ is given by
\begin{eqnarray}
\label{eq:MAPEst}
\hat{x}_{MAP} = \argmin_{x \in \mathbb{R}^N} \{ l( x) + \beta s(x)\},
\end{eqnarray}
where $l( x) = -\log p(y|x)$, $\beta s(x) = -\log p(x)$,
and $\beta$ is a positive scalar used to control the level
of regularization in the MAP reconstruction.
In order to allow for the possibility of convex constraints,
we will allow both $l(x)$ and $s(x)$ to take values on the extended real line, $\mathbb{R}\cup \{ +\infty \}$.
Using this convention, we can, for example, enforce positivity by setting $l(x) = +\infty $ for $x\leq 0$.

Splitting the variable $x$ of equation~\eqref{eq:MAPEst} results in an equivalent 
expression for the MAP estimate given by
\begin{eqnarray}
\label{eq:MAPEst_reformed}
(\hat{x}, \hat{v}) = \arg \mathop{\min_{x, v \in \mathbb{R}^N }}_{ x=v } \{l(x) + \beta s(v)\} \ .
\end{eqnarray}
This contained optimization problem can then be computed
by solving the following unconstrained augmented Lagrangian cost function given by
\begin{eqnarray}
\label{eq:Lagrangian}
L_{\lambda}(x, v; u) = l(x) + \beta s(v) + \frac{1}{2\sigma_{\lambda}^2}\|x - v + u\|_2^2 - \frac{\|u\|_2^2}{2\sigma_{\lambda}^2},
\end{eqnarray}
where $u$  must be chosen to meet the constraint of $x=v$,
and $\sigma_{\lambda} >0$ is the augmented Lagrangian parameter\footnote{ The augmented Lagrangian parameter, $\sigma_{\lambda}$, is related to the ADMM penalty parameter, $\lambda$, through a simple expression: $\sigma_{\lambda} = \frac{1}{\sqrt{\lambda}}$.}.

It is well known that the solution to equation~\eqref{eq:Lagrangian}
may be computed using the ADMM algorithm.
For this particular problem, the ADMM algorithm consists of iteration over the following steps:
\begin{eqnarray}
\hat{x} &\leftarrow& \arg \min_{x\in \mathbb{R}^N } L_{\lambda}(x, \hat{v}; u) \\
\hat{v} &\leftarrow& \arg \min_{v\in \mathbb{R}^N } L_{\lambda}(\hat{x}, v ; u) \\
u &\leftarrow& u + (\hat{x} - \hat{v}) \ .
\end{eqnarray}

In fact, if $l(x)$ and $s(x)$ are both proper, closed, and convex functions,
and a saddle point solution exists \cite{gabay1976dual, eckstein1992douglas, boyd2011distributed}, 
then it is well known that the ADMM converges to the global minimum.

We can express the ADMM iterations more compactly
by defining two operators.
The first is an inversion operator $F$ defined by
\begin{eqnarray}
\label{eq:F}
F(\tilde x; \sigma_{\lambda}) = \argmin_{x \in \mathbb{R}^N } \left\{ l(x) + \frac{\|x - \tilde x\|_2^2}{2\sigma_{\lambda}^2} \right\} \;,
\end{eqnarray}
and the second is a denoising operator $H$ given by
\begin{eqnarray}
\label{eq:H}
H(\tilde{v}; \sigma_n) = \argmin_{v \in \mathbb{R}^N} \left\{ \displaystyle\frac{\|\tilde{v}-v\|_2^2}{2\sigma_n^2} + s(v) \right\} \;,
\end{eqnarray}
where $\sigma_n = \sqrt{\beta} \sigma_{\lambda}$ has the interpretation of being
the assumed noise standard deviation in the denoising operator.
Moreover, we say that $H$ is the proximal mapping for the proper, closed, and convex function $s:\mathbb{R}^N \rightarrow \mathbb{R}\cup \{ +\infty \}$.

Using these two operators, we can easily derive the plug-and-play algorithm
shown in Algorithm~\ref{alg:PlugAndPlayAlgorithm} as an alternative form of the ADMM iterations.
This formulation has a number of practical and theoretical advantages.
First, in this form we can now ``plug in" denoising operators that are not in the 
explicit form of the optimization of equation~(\ref{eq:H}).
So for example, we will later see that popular and effective denoising operators
such as non-local means (NLM) \cite{buades2005non} or BM3D \cite{DabovBM3D07},
which are not easily represented in an optimization framework can be used in the
plug-and-play iterations.
Second, this framework allows for decomposition of the problem into separate
software systems for the implementation of the inversion operator, $F$,
and the denoising operator, $H$.
In practice, as software systems for large inversion problems become more
complex, the ability to decompose them into separate modules,
while retaining the global optimality of the solution,
can be extremely valuable. 

The plug-and-play algorithm requires the selection of two parameters, $\beta$ and $\sigma_{\lambda}$,
and then the $\sigma_n = \sqrt{\beta} \sigma_{\lambda}$.
The unit-less parameter $\beta$ can typically be chosen to be near 1,
with larger or smaller values producing more or less regularization, respectively.
In theory, the value of $\sigma_{\lambda}$ does not affect the reconstruction for a convex optimization problem, 
but in practice, a well-chosen value of $\sigma_{\lambda}$ 
can substantially speed up ADMM convergence \cite{boyd2011distributed, wahlberg2012admm, ghadimi2015optimal};
so the careful choice of $\sigma_{\lambda}$ is important.
Our approach is to choose the value of $\sigma_{\lambda}$ to be approximately equal 
to the amount of variation in the reconstruction.
Formally stated, we choose
\begin{eqnarray}
\label{eq:sigma_lambda}
\sigma_{\lambda}^2 \approx \text{var}[x|y] \ .
\end{eqnarray}
This choice for the value of $\sigma_{\lambda}^2$ is motivated by its role as the inverse regularizer in equation~\eqref{eq:F}.
In practice, this can be done by first computing an approximate reconstruction
using some baseline algorithm,
and then computing the sample variance in the approximate reconstruction.

Of course, for an arbitrary denoising algorithm,
the question remains of whether the plug-and-play algorithm converges?
The following section provides practical conditions
for the denoising operator to meet that ensure convergence
of the iterations. 

\begin{algorithm}[t]
\small
\begin{algorithmic}
\State initialize $\hat{v}$ 
\State $u \leftarrow 0$ 
\While {not converged}

\State $\tilde x \leftarrow \hat v - u $
\State $\hat{x} \leftarrow F(\tilde{x}; \sigma_{\lambda}) $
\State $\tilde{v} \leftarrow \hat{x} + u $
\State $\hat{v} \leftarrow H(\tilde{v}; \sigma_n) $
\State $u \leftarrow u + (\hat{x} - \hat{v}) $

\EndWhile

\end{algorithmic}
\caption{\small Plug-and-play algorithm for implementation of a general forward model $F( \tilde{x} ; \sigma_\lambda )$,
and a prior model specified by the denoising operator in $H( \tilde{v} ; \sigma_n )$.} 
\label{alg:PlugAndPlayAlgorithm}
\end{algorithm}

\section{Convergence of the plug-and-play algorithm}
\label{sec:Conditions}

It is well known that the ADMM algorithm is guaranteed to converge
under appropriate technical conditions.
For example, if the optimization problem is convex and a saddle point solution exists
then the iterations of ADMM converge~\cite{gabay1976dual, eckstein1992douglas, boyd2011distributed}.
However, in our plug-and-play approach,
we will be using general denoising algorithms to implement the operator $H(\tilde{v}; \sigma_n)$,
and therefore, the function $s(x)$ is not available to inspect.
This raises the question of what conditions must $H(\tilde{v}; \sigma_n)$ 
and $l(y; x)$ must satisfy in order 
to ensure that the plug-and-play algorithm converges.

In the following theorem,
we give conditions on both the log likelihood function, $l(x)$,
and the denoising operator, $H(x)$, 
that are sufficient to guarantee convergence of the plug-and-play algorithm
to the global minimum of some implicitly defined MAP cost function.
This is interesting because it does not ever require that one know or explicitly specify
the function $s(x)$. 
Instead, $s(x)$ is implicitly defined through the choice of $H(x)$.

\begin{theorem}
\label{theorem:PNP-Convergence}
Let the negative log likelihood function $l:\mathbb{R}^N \rightarrow \mathbb{R} \cup \{ +\infty \}$
and the denoising operator $H: \mathbb{R}^N \rightarrow \mathbb{R}^N$ 
meet the following conditions: 
\begin{enumerate}
\item 
$H(x )$ is a continuously differentiable function on $\mathbb{R}^N$;
\item
$\forall x\in \mathbb{R}^N$, $\nabla H(x)$ is a doubly stochastic matrix;
\item 
There exist a $y$ in the range of $H$ such that $l(y) <\infty$;
\item $l(x)$ is a proper closed convex function 
which is lower bounded by a function $f( \| x \| )$ such that 
$f(x)$ is monotone increasing with
$$
\lim_{\alpha \rightarrow \infty } \frac{ f(\alpha ) }{\alpha } = \infty \ .
$$
\end{enumerate}
Then the following results hold:
\begin{enumerate}
\item $H$ is a proximal mapping for some proper closed convex function $s(x)$;
\item There exists a MAP estimate,  $\hat{x}_{MAP}$, such that
$$
p^* = \inf_{x\in \mathbb{R}^N} \left\{ l(x) + \beta s(x) \right\} = l ( \hat{x}_{MAP} ) + \beta s ( \hat{x}_{MAP} ) \ ;
$$
\item The plug-and-play algorithm converges in the following sense,
\begin{eqnarray*}
\label{eq:ADMM_convergence}
\lim_{k\to\infty}\{\hat{x}^{(k)} - \hat{v}^{(k)}\} &=& 0;\\
\lim_{k\to\infty} \{ l(\hat{x}^{(k)}) + \beta s(\hat{v}^{(k)}) \} &=& p^*  \ ,
\end{eqnarray*}
where $\hat{x}^{(k)}$ and $\hat{v}^{(k)}$ denote the result of the $k^{th}$ iteration. 
\end{enumerate}
\end{theorem}

The proof of this theorem, which is presented in Appendix~\ref{sec:AppendixPNPProof},
depends on a powerful theorem proved by Moreau in 1965 \cite{moreau1965proximite}.
This theorem states that $H$ is a proximal mapping if and only if
it is non-expansive and the sub-gradient of a convex function on $\mathbb{R}^N$.
Intuitively, once we can show that the denoising operator, $H$,
is a proximal mapping, then we know that is effectively implementing
an update step from of the ADMM algorithm of equation~(\ref{eq:H}).

The first and second conditions of the theorem ensure that the
conditions of Moreau's theorem are met.
This is because the doubly stochastic structure of $H(x)$ ensures that 
$H$ is the gradient of some function $\phi$,
that $\phi$ is convex, and that $H$ is non-expansive.

The additional two conditions of Theorem~\ref{theorem:PNP-Convergence}
ensure that the MAP estimate actually exists for the problem.
Importantly, this is done without explicit reference to the prior function $s(x)$. 
More specifically, the third condition ensures
that the set of feasible solutions is not empty,
and the fourth condition ensures that the MAP cost
function takes on its global minimum value,
i.e., that the minimum is not achieved toward infinity. 

Importantly, in the next section, we will show that 
real denoising operators can be modified to meet the conditions of this theorem.
In particular, the symmetrized non-local means filters investigated
by Milanfar \cite{milanfar2013symmetrizing} are designed to create a symmetric gradient.

\section{Non-Local Means Denoising with Doubly Stochastic Gradient}
\label{section:SGNLM}

In order to satisfy the conditions for convergence,
the gradient of the denoising operator must be a doubly-stochastic matrix.
However, the standard NLM denoising algorithm does not satisfy this condition.
So in this section, we introduce a simple modification of the NLM approach,
which we refer to as the doubly stochastic gradient NLM (DSG-NLM),
that satisfies the required convergence conditions.
Interestingly, the symmetrized non-local means filters investigated
by Milanfar \cite{milanfar2013symmetrizing} also achieve a symmetric gradient,
but requires the use of a more complex iterative algorithm to symmetrize the operator.

The NLM algorithm is known to produce much higher quality results than traditional local smoothing-based denoising methods \cite{buades2005non}.
It works by estimating each pixel \footnote{All discussion remains valid even if we consider voxels instead of pixels.} as a weighted mean of all pixels in the image \footnote{In practice, we only compute the weighted mean of pixels in a search window, instead of the whole image.}. 
In this section, $\tilde v$ will denote a noisy image with voxel values $v_s$ at locations $s\in S$.
Generally, $S$ is a discrete lattice,
so for 2D images $S = \mathbb{Z}^2$ and for 3D volumes $S = \mathbb{Z}^3$.

Using this notation, the NLM denoising method can be represented as
\begin{eqnarray}
\label{eq:3D_NLM}
\hat{v}_s = \sum_{r \in \Omega_s } w_{s, r} \tilde v_r \ ,
\end{eqnarray}
where $\hat{v}_s$ is the denoised result,
the coefficients ${w}_{s,r}$ are the NLM weights, 
and $\Omega_s$ is the NLM search window defined by
\vspace{-2mm}
$$
\Omega_s = \left\{  r \in S : \| r - s \|_\infty \leq N_s \right\} \ .
$$
Note that the integer $N_s$ controls the size of the NLM search window.
In general, larger values of $N_s$ can yield better results
but at the cost of higher computational cost.

Using this notation, the plug-and-play denoising operator is given by
$$
H( \tilde v ; \sigma_n ) = W \tilde v  \ ,
$$
where the matrix 
$$
W_{s,r} = \left\{
\begin{array}{ll}
w_{s,r} & \mbox{if $r\in \Omega_s$} \\
0 & \mbox{otherwise}
\end{array}
\right. \ .
$$
Now if we fix the weights, then it is clear that 
$$
\nabla H ( v; \sigma_n ) = W \ .
$$
So the condition~2 of Theorem~\ref{theorem:PNP-Convergence},
simply requires that $W$ be a doubly stochastic matrix.

In fact, $W$ is guaranteed to be doubly-stochastic if it is symmetric
with positive entries and rows (or columns) that sum to 1.
The following modified procedure for computing the NLM weights
ensures that these properties hold.
We start by defining $P_s \in \mathbb{R}^{N_p^2}$ to be a 
patch of size $N_p \times N_p$ centered at position $s$.
Then we compute the weights through the following 3-step algorithm.
\begin{eqnarray}
\label{eq:DSG-NLMWeightsStep1}
w_{s,r} &\gets& \exp{\left\{ \frac{-\|P_r - P_s\|_2^2}{2 N_p^2 \sigma_n^2}\right\}} \\[10pt]
\label{eq:DSG-NLMWeightsStep2}
w_{s, r} &\gets& \frac{ w_{s, r} }{ \sqrt{ \left( \sum_{r \in \Omega_s} w_{s, r} \right) \left( \sum_{s \in \Omega_r} w_{r, s} \right) }} \\[10pt]
\label{eq:DSG-NLMWeightsStep3}
w_{s, s} &\gets& w_{s, s} - \left(\sum_{r \in \Omega_s} w_{s, r} - 1\right) \ .
\end{eqnarray}
Notice that all three steps of~equations~\eqref{eq:DSG-NLMWeightsStep1}, \eqref{eq:DSG-NLMWeightsStep2}, and~\eqref{eq:DSG-NLMWeightsStep3}
are symmetric in $s$ and $r$, so they produce symmetric weights with the property that $w_{s,r} = w_{r,s}$.
While equation~\eqref{eq:DSG-NLMWeightsStep2} results in rows and columns that are approximately normalized,
and equation~\eqref{eq:DSG-NLMWeightsStep3} guarantees normalization by adjusting the diagonal coefficient of the matrix $W$.
Theoretically, equation~\eqref{eq:DSG-NLMWeightsStep3} could produce a negative coefficient, 
but in practice this does not occur in real data for two reasons.
First, the diagonal coefficient, $w_{s,s}$ is always the largest value generated 
in step~1 of~equation~\eqref{eq:DSG-NLMWeightsStep1} because $\| P_s - P_s \|_2^2 = 0$.
Second, the normalization of equation~\eqref{eq:DSG-NLMWeightsStep2} typically makes the subtracted quantity of~equation~\eqref{eq:DSG-NLMWeightsStep3} small.

Therefore, this algorithm generates a matrix $W$ which is symmetric with rows and columns that sum to 1,
and in all practical cases, non-negative elements.
This makes $W$ a doubly stochastic matrix,
so it fulfills condition~2 of Theorem~\ref{theorem:PNP-Convergence}
as is required for guaranteed convergence of the plug-and-play algorithm.


\section{3D Bright Field EM Forward Model}
\label{sec:F_operator}

In this section, we formulate the explicit form of the inversion operator, $F( x, \sigma_\lambda )$,
for the application of 3D bright field EM tomography.
For this problem, we adopted both the forward model and optimization algorithms described in \cite{venkatakrishnan2014TCI}.
More specifically, the negative log likelihood function is given by
\begin{align}
\nonumber
l(x, d, \sigma ) = & \frac{1}{2} \sum_{k=1}^K \sum_{i=1}^M \beta_{T, \delta}\left((y_{k, i} - A_{k, i, *}x - d_k) \frac{\sqrt{\Lambda_{k, ii}}}{\sigma}\right)\\
&+ MK\log{(\sigma)} + C\ ,
\nonumber
\end{align}
where $K$ is the number of tilts, 
$\lambda_{k, i}$ is the electron counts corresponding to the $i$-th measurement at the $k$-th tilt, $y_{k, i} = -\log{\lambda_{k, i}}$, 
$\lambda_{D, k}$ is the blank scan value at the $k$-th tilt, $d_k = -\log{\lambda_{D, k}}$, 
$A_k$ is the $M\times N$ tomographic forward projection matrix associated with the $k$-th tilt, 
$A_{k, i, *}$ is the $i$-th row of $A_k$, $\sigma^2$ is a proportionality constant, 
$\Lambda_k$ is a diagonal matrix whose entries are set such that $\frac{\sigma^2}{\Lambda_{k, ii}}$ is the variance 
of $y_{k, i}$, $d = [d_1, ..., d_K]$ is the offset parameter vector, 
$C$ is a constant, 
and $\beta_{T, \delta}(\cdot)$ is the generalized Huber function defined as,
\begin{eqnarray}
\label{eq:Huber_function}
 \beta_{T, \delta}(x) =
  \begin{cases} 
      \hfill x^2    \hfill & \text{ if $|x| < T$} \\
      \hfill 2\delta T|x| + T^2(1-2\delta) \hfill & \text{ if $|x| \geq T$} \ .
  \end{cases}
\end{eqnarray}
The generalized Huber function is used to reject measurements with large errors.
This is useful because measurement may vary from the assumed model 
for many practical reasons.
For example, in bright field EM, Bragg scatter can cause highly attenuated measurements
that otherwise would cause visible streaks on the reconstruction \cite{de2003introduction}.

To compute the inversion operator $F$ of equation~\eqref{eq:F}, 
we minimize the cost function below with respect to $x$, $d$, and $\sigma$.
\begin{align}
\label{eq:augmented_cost_function}
&c(x, d, \sigma; \tilde{x}, \sigma_{\lambda}) \nonumber \\
&= \frac{1}{2} \sum_{k=1}^K \sum_{i=1}^M \beta_{T, \delta}\left((y_{k, i} - A_{k, i, *}x - d_k)\displaystyle\frac{\sqrt{\Lambda_{k, ii}}}{\sigma}\right) \nonumber \\
&+ MK\log{(\sigma)} + \displaystyle\frac{\|x - \tilde{x}\|_2^2}{2\sigma_{\lambda}^2}.
\end{align}
So the inversion operator is computed as
\begin{eqnarray}
\label{eq:BFMBIR-inversion-operator}
F( \tilde{x} ; \sigma_\lambda ) = \arg \min_{x \geq 0, d, \sigma} c(x, d, \sigma; \tilde{x}, \sigma_{\lambda}) .
\end{eqnarray}
As in the case of sparse interpolation, we set $c(x, d, \sigma; \tilde{x}, \sigma_\lambda ) = +\infty $ for $x<0$ in order to enforce positivity.

The details of the optimization algorithm required for equation~(\ref{eq:BFMBIR-inversion-operator})
are described in \cite{venkatakrishnan2014TCI}.
The optimization algorithm is based on alternating minimization
with respect the the three quantities
and it uses a majorization based on a surrogate function to handle the minimization
of the generalized Huber function \cite{stevenson1990fitting}.

For this complex problem,
we note some practical deviations from the theory.
First, the negative log likelihood function, $l(x)$, is not convex in this case,
so the assumptions of the plug-and-play convergence do not hold.
Moreover, with such a non-convex optimization, it is not possible to guarantee
convergence to a global minimum, but in practice most optimization algorithms
generate very good results.
In addition, this cost function also violates condition~4 of Theorem~\ref{theorem:PNP-Convergence}
because it only grows at a linear rate as $\| x \|\rightarrow +\infty $.
Again, this condition is used to guarantee that the plug-and-play algorithm
does not drift off to a minimum tending to infinity.
However, in practice, we have never observed this to happen with real data sets
and useful denoising operators.
Finally, the global optimization of equation~(\ref{eq:BFMBIR-inversion-operator})
is approximated by three iterations of alternating minimization with respect
to $x$, $d$, and $\sigma$. 
Nonetheless, in our experimental results section,
we will illustrate our empirical observation that the plug-and-play algorithm consistently converges 
even with these approximations to the ideal case.


\section{Sparse Interpolation Forward Model}
\label{section:Interpolation}

In this section, we formulate the explicit form of the inversion operator, $F( x, \sigma_\lambda )$,
for the application of sparse interpolation.
More specifically, our objective will be to recover and image $x\in \mathbb{R}^N$
from a noisy and sparsely subsampled version denoted by $y\in \mathbb{R}^M$
where $M<<N$.
More formally, the forward model for this problem is given by
\begin{equation}
\label{eq:sparse_interpolation}
y = Ax + \epsilon \ ,
\end{equation}
where $A\in \mathbb{R}^M \times \mathbb{R}^N$ matrix.
Each entry $A_{i,j}$ is either 1 or 0 depending on if the $j^{th}$ pixel is taken
as the $i^{th}$ measurement.
Also, each row of $A$ has exactly one non-zero entry,
and each column of $A$ may either be empty or have one non-zero entry.
We also define $I(j) = \sum_{i} A_{i,j}$
so that $I(j) = 1$ when the $j^{th}$ pixel is sampled,
and $I(j) = 0$, if it is not.
Furthermore, $\epsilon$ is an $M$-dimensional vector 
of i.i.d.\ Gaussian random variables with mean zero and variance $\sigma_w^2$.

For such a sparse sampling system,
we can write the negative log likelihood function as
\begin{equation}
l(x) = \frac{ 1 }{ 2 \sigma_w^2 } \| y - A x \|_2^2 + C \ ,
\end{equation}
where $C$ is a constant.
In order to enforce positivity, 
we also modify the negative likelihood function by setting $l(x) = +\infty$ for $x<0$.
We include positivity in $l(x)$ rather than in the denoising operator so that $H$ remains
continuously differentiable.

\noindent Using equation~(\ref{eq:F}), the interpolation inversion operator is given by
$$
F( \tilde{x} ; \sigma_{\lambda}) = \argmin_{x \geq 0} \left\{ \frac{1}{2 \sigma_w^2} \|y-Ax\|_2^2 + \frac{1}{2\sigma_{\lambda}^2} \|x - \tilde x\|_2^2 \right\}.
$$
Due to the simple structure of the matrix $A$,
we can also calculate an explicit pixel-wise expression for $F$.
Moreover, if we let $\sigma_w^2 = 0$, then $F$ reduces to the following form
\begin{equation}
 F_i(\tilde x; \sigma_{\lambda}) =
  \begin{cases} 
      \hfill \left[ y_i \right]_+   \hfill & \text{ if $I(i) = 1$} \\
      \hfill \left[ \tilde x_i \right]_+ \hfill & \text{ if $I(i) = 0$} \\
  \end{cases}.
\end{equation}
where $[\cdot]_+$ represents zeroing of any negative argument.
In this case, the interpolation is forced to take on the measured
values at the sample points.


\section{Results}
\label{section:Results}

In this section, we present experimental results on both real and simulated data
for the applications of bright-field EM tomography and sparse interpolation.
For all experiments, we present convergence plots that compare both primal and dual residual convergence resulting from using different priors.
The normalized primal and dual residues \cite[p. 18]{boyd2011distributed}, $r^{(k)}$ and $s^{(k)}$ respectively, at the $k$-th iteration of the P\&P algorithm are given by
\begin{eqnarray}
r^{(k)} = \displaystyle\frac{\| \hat{x}^{(k)} - \hat{v}^{(k)} \|_2}{\| \hat{x}^{(\infty)} \|_2};\\
s^{(k)} = \displaystyle\frac{\| \hat{v}^{(k)} - \hat{v}^{(k-1)} \|_2}{\| u^{(k)} \|_2},
\end{eqnarray}
where $\hat{x}^{(k)}$, $\hat{v}^{(k)}$, and $u^{(k)}$ are the values of $\hat{x}$, $\hat{v}$, and $u$ respectively after the $k$-th iteration of the plug-and-play algorithm, respectively, and $\hat{x}^{(\infty)}$ is the final value of the reconstruction, $\hat{x}$.

\vspace{-5mm}

\subsection{Bright Field EM Tomography}
\label{subsec:tomography_results}

In this section, we present the results of bright field tomographic reconstruction of (1) a simulated dataset of aluminum spheres of varying radii, (2) a real dataset of aluminum spheres, and (3) a real dataset of silicon dioxide.
We compare four reconstruction methods -- filtered backprojection, MBIR with qGGMRF prior \cite{thibault2007three}, plug-and-play reconstructions with 3D NLM and 3D DSG-NLM as prior models.
We used qGGMRF, 3D NLM and 3D DSG-NLM as prior models within the plug-and-play framework. Filtered backprojection was used as the initialization for all MBIR-based reconstructions.
All the reconstruction results shown below are $x$-$z$ slices (i.e., slices parallel to the electron beam).
The qGGMRF parameters used for all reconstructions are as follows: $q = 1$, $p = 1.2$, and $c = 0.001$. The NLM and DSG-NLM patch size used for all reconstructions is $5 \times 5 \times 5$. In order to meet the conditions of convergence, we stopped adapting the DSG-NLM weights at 20 iterations of the plug-and-play algorithm. The P\&P parameters used are given in Table~\ref{table:tomo_parameters_table}.

In all the experiments, we observe from Tables~\ref{table:primal_convergence_error_tomography} and \ref{table:dual_convergence_error_tomography} that the DSG-NLM ensures that the plug-and-play algorithm converges fully, while NLM achieves convergence to within a fraction of a percent.

\subsubsection{Aluminum spheres (simulated) dataset}

The aluminum spheres simulated dataset contains 47 equally-spaced tilts about the $y$-axis, spanning $[-70\degree, +70\degree]$.
The attenuation co-efficient of the spheres are assumed to be $7.45 \times 10^{-3}$ nm.
The noise model is Gaussian, with variance set equal to the mean. The phantom also contains effects that resemble Bragg scatter. 
The dimensions of the phantom are 256 nm, 512 nm, and 512 nm -- along $z$, $x$, and $y$ axes, respectively.

\begin{figure}[h]
\center
\includegraphics[scale=0.35]{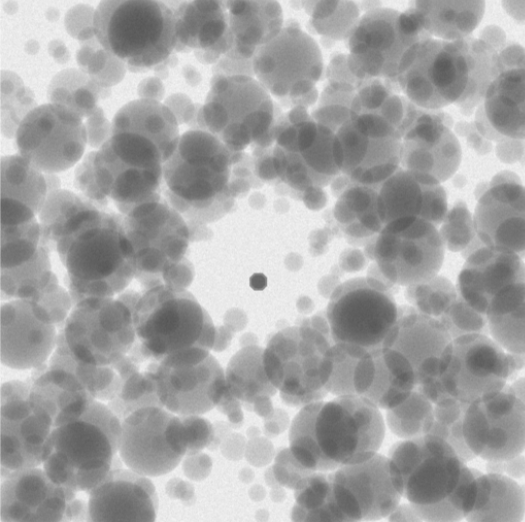}
\caption{0$\degree$ tilt of the aluminum spheres (simulated) dataset.}
\label{fig:phantom}
\end{figure}

\begin{figure}[H]
\centering
\subfigure[The aluminum spheres phantom (ground truth)]{%
\includegraphics[width = 6cm]{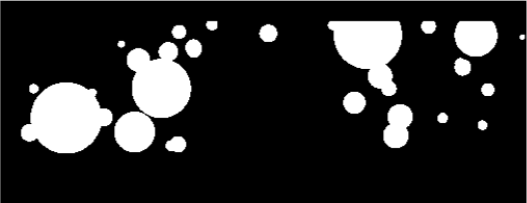}}
\quad
\subfigure[Filtered Backprojection]{%
\includegraphics[width = 6cm]{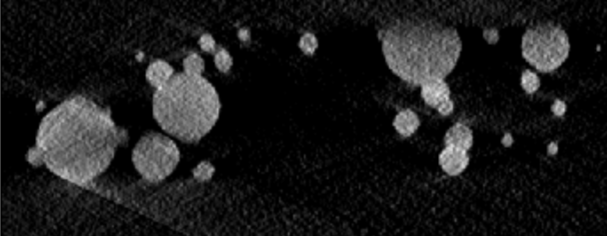}}
\quad
\subfigure[qGGMRF ($T = 3$; $\delta = 0.5$)]{%
\includegraphics[width = 6cm]{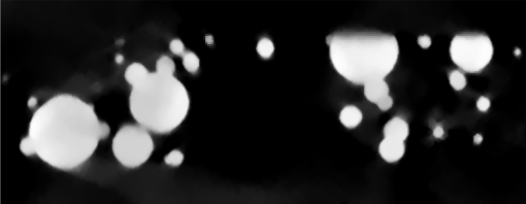}}
\quad
\subfigure[3D NLM using plug-and-play]{%
\includegraphics[width = 6cm]{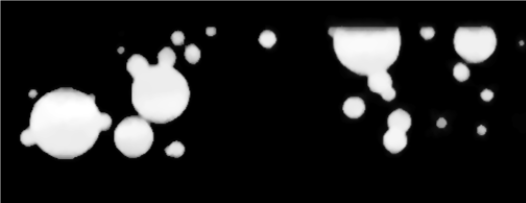}}
\quad
\subfigure[3D DSG-NLM using plug-and-play]{%
\includegraphics[width = 6cm]{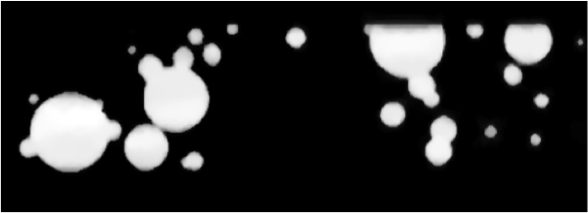}}
\caption{Tomographic reconstruction of the simulated aluminum spheres dataset. 
NLM and DSG-NLM reconstructions are clearer and relatively artifact-free.}
\label{fig:Exp1_results}
\end{figure}

Fig. \ref{fig:phantom} shows a $0\degree$ tilt projection of the simulated TEM data. 
Since this is a bright-field image, the aluminum spheres appear dark against a bright background. Fig. \ref{fig:Exp1_results} shows the ground truth along with three reconstructions of slice 280 along the $x$-$z$ plane.
The NLM and DSG-NLM reconstructions have no shadow artifacts, 
and also have low RMSE values (see Table~\ref{table:Sim_Al_RMSE}).
The edges are also sharper in the NLM and DSG-NLM reconstructions.

\begin{table}[!htbp]
\caption{RMSE of the reconstructed Al spheres image compared to the ground truth 
\\(after 200 P\&P iterations)} 
\centering 
\begin{tabular}{|c|c|c|c|} 
\hline
\textbf{FBP}  & \textbf{qGGMRF} & \textbf{NLM} & \textbf{DSG-NLM} \\ [0.5ex]
\hline\hline
14.608 & 4.581 & 2.531 & 2.529 \\ 
$\times 10^{-4}$ nm$^{-1}$ & $\times 10^{-4}$ nm$^{-1}$ & $\times 10^{-4}$ nm$^{-1}$ & $\times 10^{-4}$ nm$^{-1}$ \\ [1ex] 
\hline 
\end{tabular}
\label{table:Sim_Al_RMSE} 
\end{table}

\vspace{1cm}

\begin{figure}[!htbp]
\centering
\includegraphics[scale=0.21]{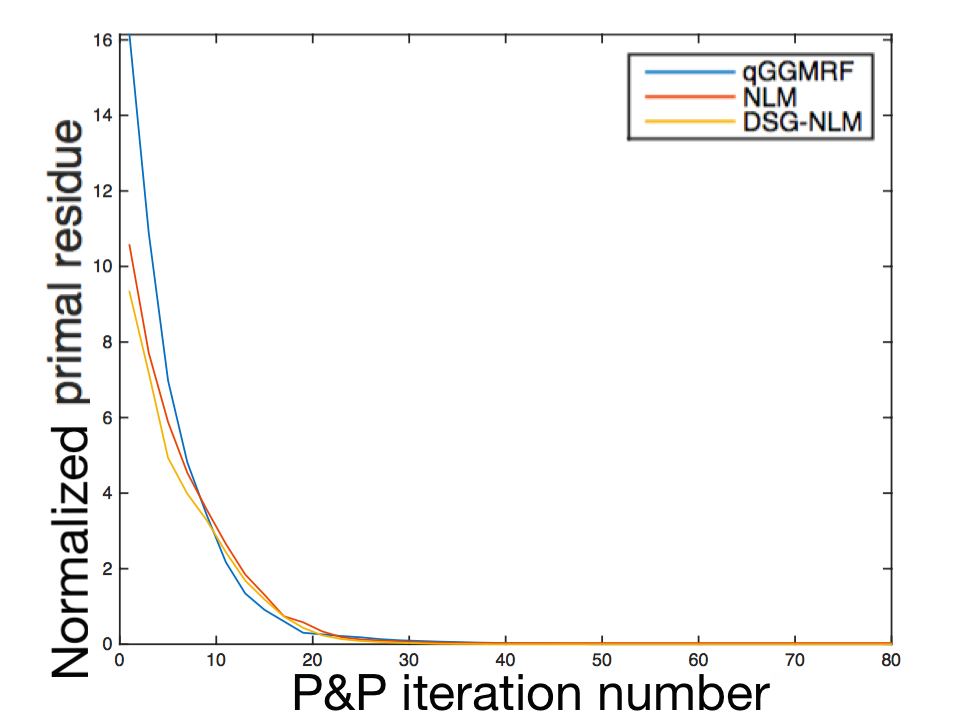}
\qquad
\includegraphics[scale=0.21]{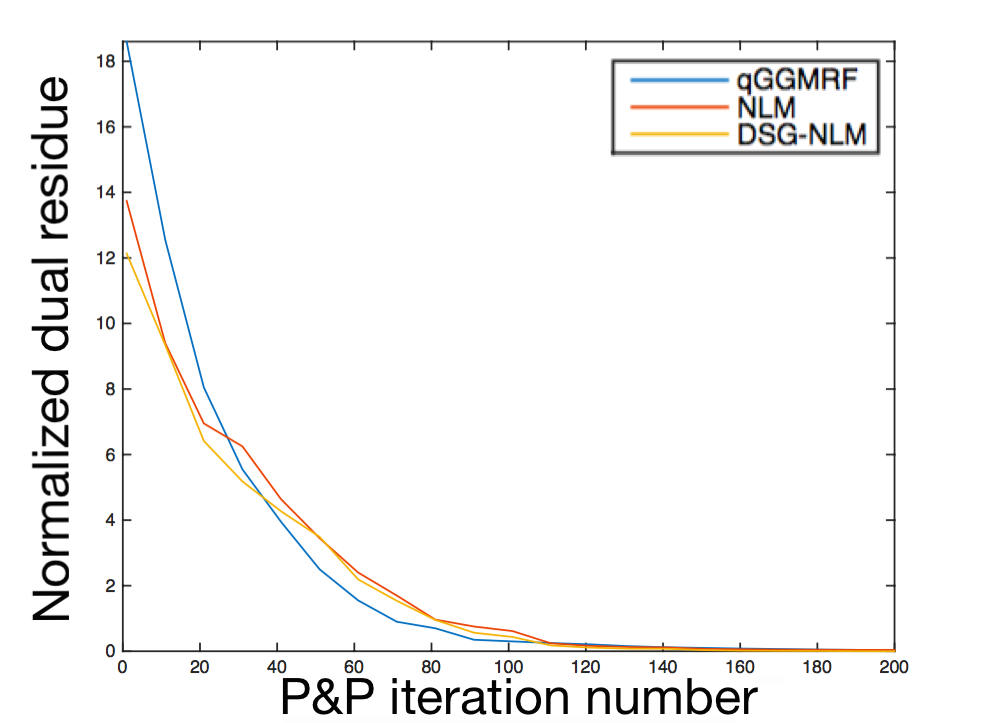}
\caption{Plug-and-play primal and dual residual convergence for tomographic reconstruction of (simulated) aluminum spheres. DSG-NLM achieves complete convergence.}
\label{fig:tomography_convergence}
\end{figure}

\subsubsection{Aluminum spheres (real) dataset}

\begin{figure}[h]
\center
\includegraphics[scale=0.4]{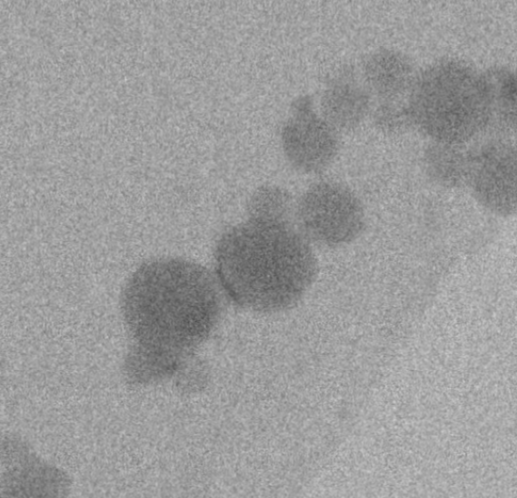}
\caption{0$\degree$ tilt of the very noisy aluminum spheres (real) dataset.}
\label{fig:Al_real_data}
\end{figure}

\begin{figure}[H]
\centering
\subfigure[Filtered Backprojection]{%
\includegraphics[width = 6.5cm]{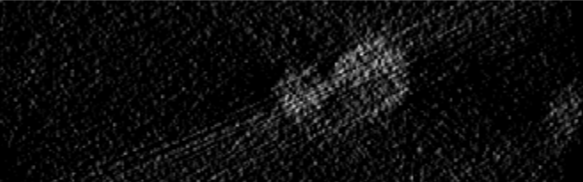}}
\quad
\subfigure[qGGMRF ($T = 3$; $\delta = 0.5$)]{%
\includegraphics[width = 6.5cm]{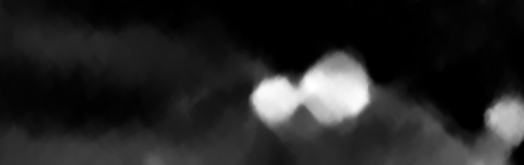}}
\quad
\subfigure[3D NLM using plug-and-play]{%
\includegraphics[width = 6.5cm]{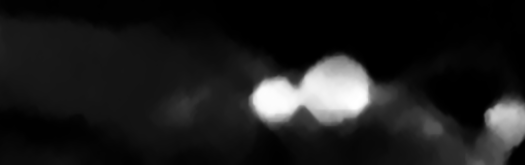}}
\quad
\subfigure[3D DSG-NLM using plug-and-play]{%
\includegraphics[width = 6.5cm]{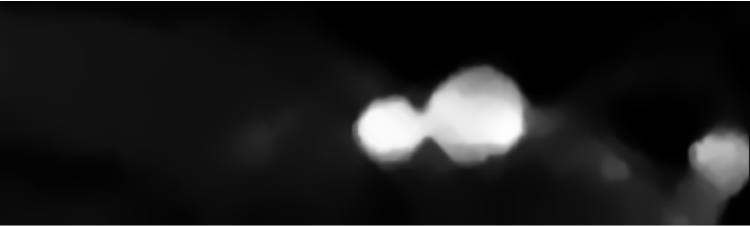}}
\caption{Tomographic reconstruction of the real aluminum spheres dataset. 
NLM and DSG-NLM reconstructions are clearer and have less smear and missing-wedge artifacts.}
\label{fig:Exp2_results}
\end{figure}

The aluminum spheres dataset (see Fig. \ref{fig:Al_real_data}) has 67 equally-spaced tilts about the $y$-axis, spanning $[-65\degree, +65\degree]$. 
Fig. \ref{fig:Al_real_data} shows a $0\degree$ tilt projection of the real aluminum spheres TEM data. Fig. \ref{fig:Exp2_results} shows three reconstructions along the $x$-$z$ plane.
The NLM-based reconstruction has less smear artifacts than the qGGMRF reconstruction, and more clarity than the filtered backprojection reconstruction. Also, the NLM and DSG-NLM reconstructions have visibly suppressed missing-wedge artifact.

\begin{figure}[!htbp]
\centering

\includegraphics[scale=0.2]{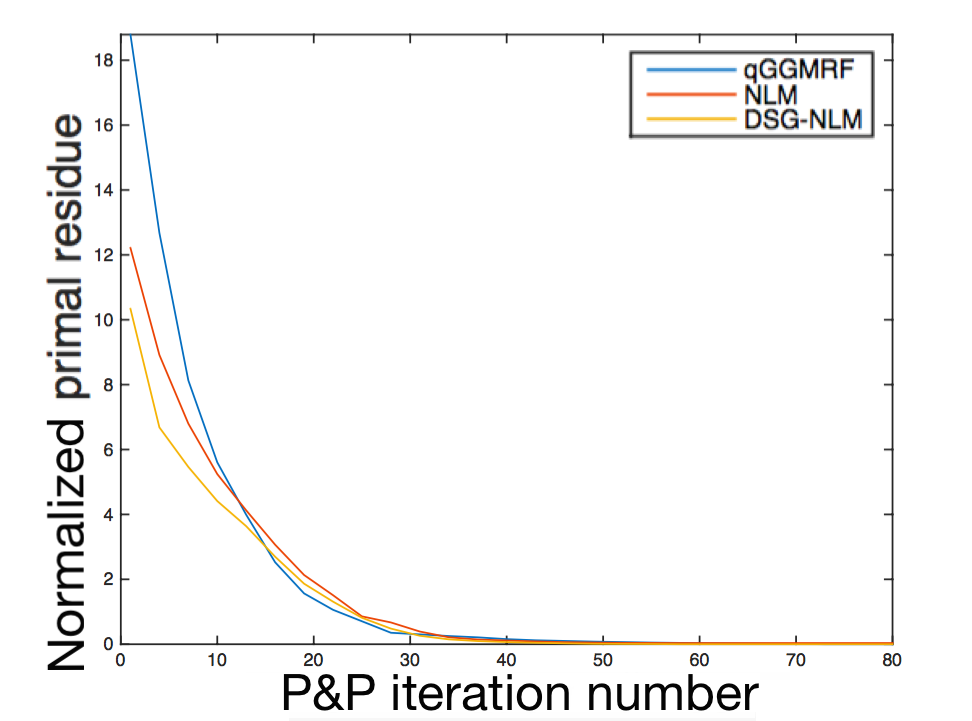}
\qquad
\includegraphics[scale=0.2]{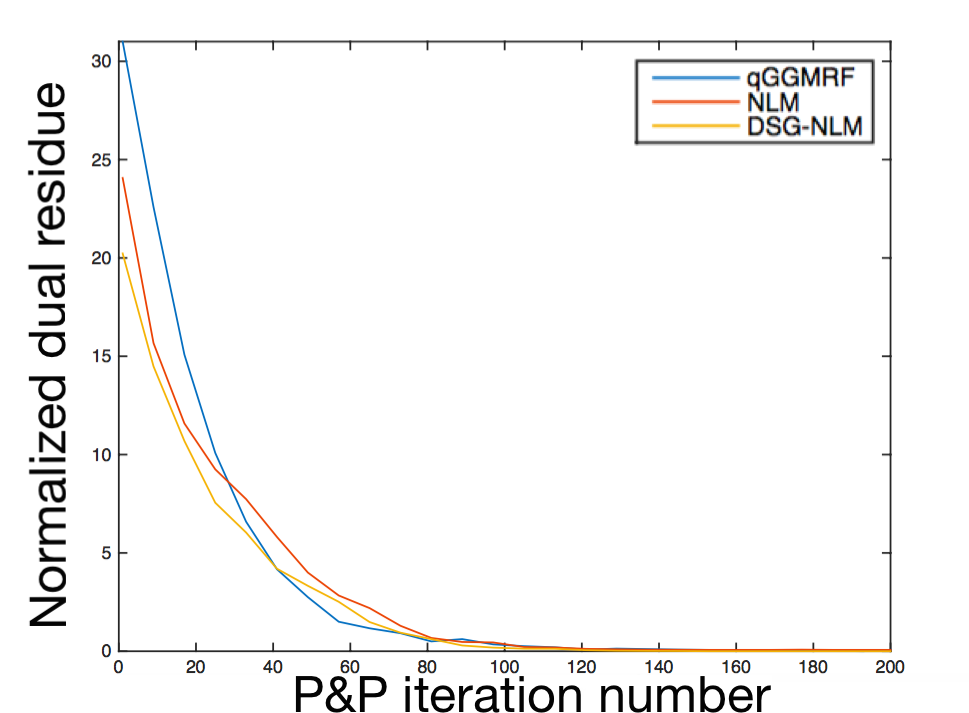}
\caption{Plug-and-play primal and dual residual convergence for tomographic reconstruction of (real) aluminum spheres. DSG-NLM achieves complete convergence.}
\label{fig:tomography_convergence}
\end{figure}

\subsubsection{Silicon dioxide (real) dataset}

\begin{figure}[h]
\center
\includegraphics[scale=0.5]{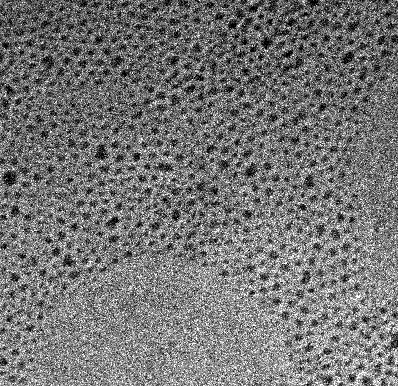}
\caption{Contrast-adjusted version of the 0$\degree$ tilt of the silicon dioxide (real) dataset.}
\label{fig:SiD_real_data}
\end{figure}

The silicon dioxide dataset (see Fig. \ref{fig:SiD_real_data}) has 31 tilts about the $y$-axis, spanning $[-65\degree, +65\degree]$.

\begin{figure}[!htbp]
\centering
\subfigure[Filtered Backprojection]{%
\adjustbox{trim={.0\width} {.0\height} {0.2\width} {.0\height},clip}%
{\includegraphics[width = 10.5cm]{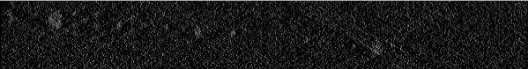}}}
\quad
\subfigure[qGGMRF ($T = 3$; $\delta = 0.5$)]{%
\adjustbox{trim={.0\width} {.0\height} {0.2\width} {.0\height},clip}%
{\includegraphics[width = 10.5cm]{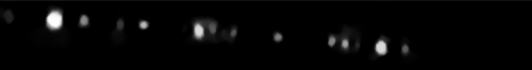}}}
\quad
\subfigure[3D NLM using plug-and-play]{%
\adjustbox{trim={.0\width} {.0\height} {0.2\width} {.0\height},clip}%
{\includegraphics[width = 10.5cm]{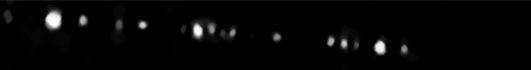}}}
\quad
\subfigure[3D DSG-NLM using plug-and-play]{%
\adjustbox{trim={.0\width} {.0\height} {0.2\width} {.0\height},clip}%
{\includegraphics[width = 10.5cm]{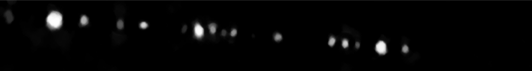}}}
\caption{Tomographic reconstruction of the silicon dioxide dataset. 
NLM reconstruction is clearer and has less smear artifacts. DSG-NLM reconstruction improves upon the NLM result through clear reconstruction of the structure on the left. }
\label{fig:Exp3_results}
\end{figure}

Fig.~\ref{fig:SiD_real_data} shows a $0\degree$ tilt projection of the real silicon dioxide TEM data. Fig. \ref{fig:Exp3_results} shows three reconstructions along the $x$-$z$ plane.
The NLM and DSG-NLM reconstructions have less smear artifacts than the qGGMRF reconstruction, and far more clarity than the filtered backprojection reconstruction.

\begin{figure}[!htbp]
\centering
\includegraphics[scale=0.21]{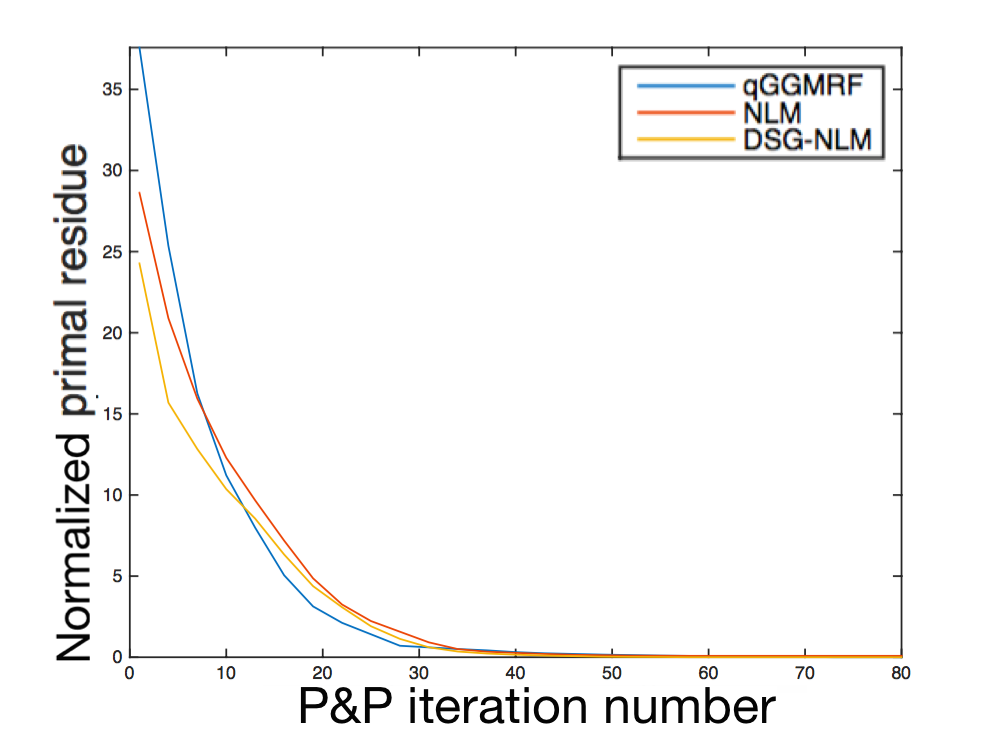}
\qquad
\includegraphics[scale=0.21]{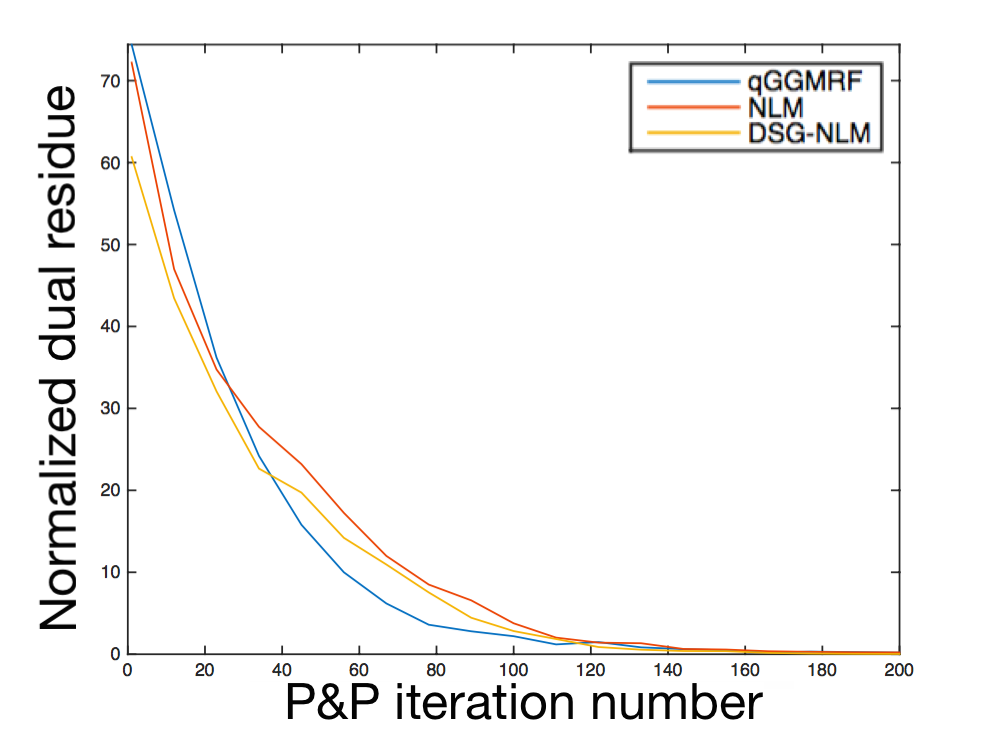}
\caption{Plug-and-play primal and dual residual convergence for tomographic reconstruction of (real) silicon dioxide. DSG-NLM achieves complete convergence.}
\label{fig:tomography_convergence}
\end{figure}

\begin{table}[h]
\caption{Plug-and-play parameters for tomographic reconstructions} 
\centering 
\begin{tabular}{|c | c | c | c|} 
\hline
 & \textbf{Al spheres} & \textbf{Al spheres} & \textbf{Silicon dioxide} \\
 & (simulated) & (real) & (real)
 \\ [0.5ex] 
\hline\hline
& & & \\
$\sigma_{\lambda}$ (nm$^{-1}$)& 8.66$\times10^{-4}$ & 8.66$\times10^{-4}$ & 8.66$\times10^{-4}$ \\ 
\hline
& & & \\
 $\beta$ & 3.68 & 4.77 & 4.25 \\
 [1ex] 
\hline 
\end{tabular}
\label{table:tomo_parameters_table} 
\end{table}

\begin{table}[!htbp]
\caption{Normalized primal residual convergence error for tomography experiments \\(after 200 P\&P iterations)}
\centering 
\begin{tabular}{| c | c | c | c |} 
\hline
\textbf{Dataset} & \textbf{qGGMRF} & \textbf{NLM} & \textbf{DSG-NLM} \\ [0.5ex] 
\hline\hline
Al spheres & $3.46\times 10^{-12}$ & $2.12\times 10^{-3}$ & $2.91\times 10^{-10}$ \\
(simulated) & & & \\
\hline
Al spheres & $7.06\times 10^{-11}$ & $3.66\times 10^{-4}$ & $6.89\times 10^{-9}$ \\
(real) & & & \\
\hline
Silicon dioxide & $4.99\times 10^{-12}$ & $8.12 \times 10^{-3}$ & $4.42\times 10^{-9}$ \\
(real) & & & \\
 [1ex] 
\hline 
\end{tabular}
\label{table:primal_convergence_error_tomography}
\end{table}

\begin{table}[!htbp]
\caption{Normalized dual residual convergence error for tomography experiments \\(after 200 P\&P iterations)}
\centering 
\begin{tabular}{| c | c | c | c |}
\hline
\textbf{Dataset} & \textbf{qGGMRF} & \textbf{NLM} & \textbf{DSG-NLM} \\ [0.5ex] %
\hline\hline
Al spheres & $1.55\times 10^{-10}$ & $7.22\times 10^{-3}$ & $8.83\times 10^{-9}$ \\
(simulated) & & & \\
\hline
Al spheres & $2.61\times 10^{-10}$ & $1.12\times 10^{-3}$ & $3.39\times 10^{-8}$ \\
(real) & & & \\
\hline
Silicon dioxide & $9.06\times 10^{-11}$ & $5.49 \times 10^{-2}$ & $5.04\times 10^{-8}$ \\
(real) & & & \\
 [1ex] 
\hline 
\end{tabular}
\label{table:dual_convergence_error_tomography}
\end{table}

\clearpage


\subsection{Sparse Interpolation}
\label{subsec:sparse_interpolation_results}

In this section, we present sparse interpolation results on both simulated and real microscope images. We show that a variety of denoising algorithms like NLM, DSG-NLM, and BM3D can be plugged in as prior models to reconstruct images from sparse samples. In all the sparse interpolation experiments, we stopped adapting the weights of the DSG-NLM after 12 iterations of the plug-and-play algorithm. The P\&P parameters used are given in Table~\ref{table:interpolation_parameters}.

Our first dataset is a set of simulated super ellipses that mimic the shapes of several material grains like Ni-Cr-Al alloy \cite{HuixiZhao1}.
The next dataset is a real microscope image of zinc oxide nano-rods \cite{koerner2009zno}.
All the images are scaled to the range $[0, 255]$.

In all experiments, the plug-and-play sparse interpolation results are clearer than Shepard interpolation results. We observe from Table \ref{table:interpolation_error} that DSG-NLM typically results in the least RMS interpolation error. The RMSE values are normalized as
$\displaystyle \frac{\| x - \hat{x} \|_2}{\| x \|_2}$, where $\hat{x}$ is the interpolated image and $x$ is the ground truth image.
Furthermore, we can see from Tables~\ref{table:primal_convergence_error_interpolation} and \ref{table:dual_convergence_error_interpolation} that DSG-NLM makes plug-and-play converge fully.

\begin{table}[!htbp]
\centering
\caption{Plug-and-play parameter, $\beta$, for the 10\% sampling case.}
\begin{center}
\begin{tabular}{|l|c|c|c|}
\hline
\textbf{Image} & \textbf{NLM} & \textbf{DSG-NLM} & \textbf{BM3D} \\
\hline\hline
& & & \\
Super ellipses & 0.9 & 0.79 & 0.55 \\
\hline
& & & \\
Zinc oxide nano-rods & 0.81 & 0.74 & 0.49 \\
\hline
\end{tabular}
\end{center}
\label{table:interpolation_parameters}
\end{table}


\begin{table}[!htbp]
\centering
\caption{Normalized primal residual convergence error for the 10\% sampling case \\(after 150 P\&P iterations)} 
\begin{tabular}{|l|c|c|c|} 
\hline
\textbf{Image} & \textbf{NLM} & \textbf{DSG-NLM} & \textbf{BM3D} \\ [0.5ex] 
\hline\hline 
& & & \\
Super & $1.31\times 10^{-3}$ & $5.41\times 10^{-8}$ & $1.20\times 10^{-3}$ \\
ellipses & & & \\
\hline
& & & \\
Zinc & $2.02\times 10^{-3}$ & $3.64\times 10^{-9}$ & $1.14\times 10^{-3}$ \\
oxide & & & \\
nano-rods & & & \\
[1ex] 
\hline 
\end{tabular}
\label{table:primal_convergence_error_interpolation} 
\end{table}

\begin{table}[!htbp]
\centering
\caption{Normalized dual residual convergence error for the 10\% sampling case 
\\(after 150 P\&P iterations)} 
\begin{tabular}{|l|c|c|c|} 
\hline
\textbf{Image} & \textbf{NLM} & \textbf{DSG-NLM} & \textbf{BM3D} \\ [0.5ex] 
\hline\hline 
& & & \\
Super & $9.10\times 10^{-3}$ & $3.58\times 10^{-7}$ & $8.71\times 10^{-3}$ \\
ellipses & & & \\
\hline
& & & \\
Zinc & $1.14\times 10^{-2}$ & $6.33\times 10^{-8}$ & $3.23\times 10^{-2}$ \\
oxide & & & \\
nano-rods & & & \\
[1ex] 
\hline 
\end{tabular}
\label{table:dual_convergence_error_interpolation} 
\end{table}

\begin{table}[!htbp]
\caption{Interpolation error (after 150 P\&P iterations): normalized RMSE of the interpolated image compared to the ground truth} 
\centering 
\begin{tabular}{|c|c|cc|} 
\hline
 & & \textbf{5\%} & \textbf{10\%} \\  
\textbf{Image}  & \textbf{Method}  & \textbf{random} & \textbf{random} \\
 & & \textbf{sampling} & \textbf{sampling} \\ [0.5ex]
\hline\hline
& & & \\
Super ellipses & Shepard & 10.61\% & 8.99\% \\
& NLM & 8.51\% & 7.12\% \\
& DSG-NLM & \textbf{8.33\%} & \textbf{6.98\%} \\
& BM3D & 9.75\% & 7.46\% \\
\hline
& & & \\
Zinc oxide nano-rods & Shepard & 6.01\% & 5.49\% \\
& NLM & 4.35\% & 3.67\% \\
& DSG-NLM & \textbf{4.18\%} & \textbf{3.39\%} \\
& BM3D & 4.72\% & 3.80\% \\
[1ex] 
\hline 
\end{tabular}
\label{table:interpolation_error} 
\end{table}

\section{Conclusions}
\label{section:Conclusions}

Microscope images of material and biological samples contain several repeating 
structures at distant locations.
High quality reconstruction of these samples is possible by exploiting 
non-local repetitive structures.
Though model-based iterative reconstruction (MBIR) could in principle 
exploit these repetitions, practically choosing the appropriate log probability term 
is very challenging.
To solve this problem, we presented the ``plug-and-play'' (P\&P) framework 
which is based on ADMM. 
ADMM is a popular method to decouple the log likelihood and 
the log prior probability terms in the MBIR cost function.
Plug-and-play takes ADMM one step further by replacing the 
optimization step related to the prior model by a denoising operation.
This approach has two major advantages:
First, it allows the use of a variety of modern denoising operators as implicit prior models;
and second, it allows for more modular implementation of software systems 
for the solution of complex inverse problems.

We next presented and proved theoretical conditions for convergence of the plug-and-play algorithm
which depend on the gradient of the denoising operator being a doubly stochastic matrix.
We also re-designed the non-local means (NLM) denoising algorithm
to have a doubly stochastic gradient, thereby ensuring plug-and-play convergence.

In order to demonstrate the value of our method,
we applied the plug-and-play algorithm to two important problems: 
bright field electron tomography and sparse image interpolation.
The results indicate that the plug-and-play algorithm when used with the NLM and DSG-NLM priors 
were able to reduce artifacts, improve clarity, and 
reduce RMSE (for the simulated dataset)
as compared to the filtered back-projection and qGGMRF reconstructions.
Then we performed sparse interpolation on simulated and real microscope images 
with as little as 5\% of the pixels sampled --
using three denoising operators: NLM, 
doubly-stochastic gradient NLM (DSG-NLM), and BM3D. 
We then compared the results against Shepard's interpolation as the baseline.
In all experiments, DSG-NLM resulted in the least RMSE and also complete convergence of the plug-and-play algorithm, as predicted by theory.


\begin{figure*}[!htbp]
\centering

\subfigure[$5\%$ sampling]{%
\includegraphics[width=2.5cm]{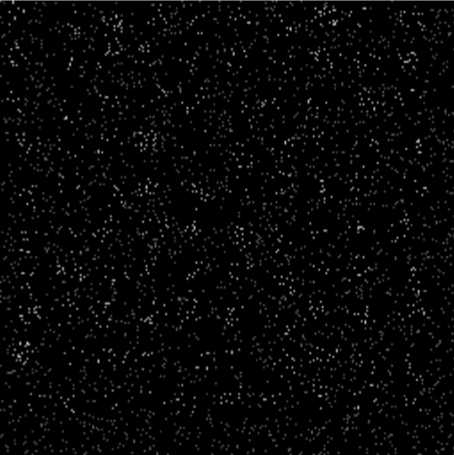}}
\quad
\subfigure[Shepard $5\%$ random sampling]{%
\includegraphics[width=2.5cm]{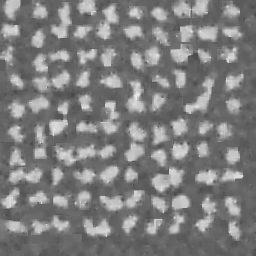}}
\quad
\subfigure[NLM $5\%$ random sampling]{%
\includegraphics[width=2.5cm]{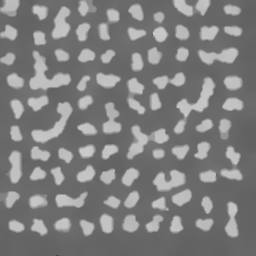}}
\quad
\subfigure[DSG-NLM $5\%$ random sampling]{%
\includegraphics[width=2.5cm]{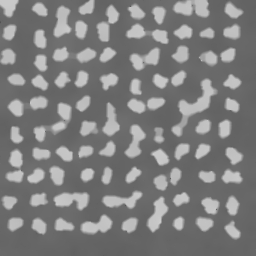}}
\quad
\subfigure[BM3D $5\%$ random sampling]{%
\includegraphics[width=2.5cm]{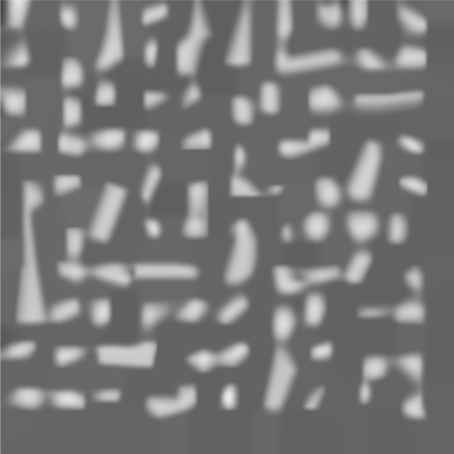}}

\subfigure[$10\%$ sampling]{%
\includegraphics[width=2.5cm]{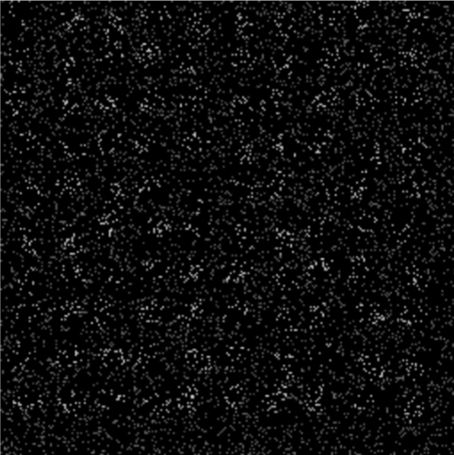}}
\quad
\subfigure[Shepard $10\%$]{%
\includegraphics[width=2.5cm]{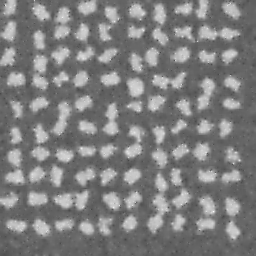}}
\quad
\subfigure[NLM $10\%$]{%
\includegraphics[width=2.5cm]{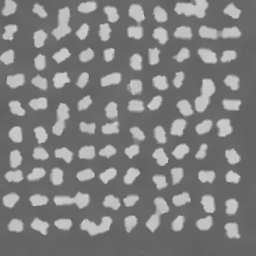}}
\quad
\subfigure[DSG-NLM $10\%$]{%
\includegraphics[width=2.5cm]{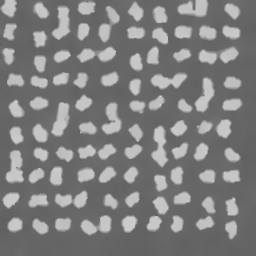}}
\quad
\subfigure[BM3D $10\%$]{%
\includegraphics[width=2.5cm]{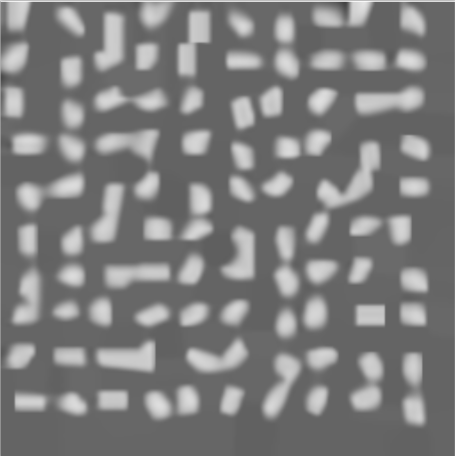}}

\subfigure[Ground truth]{%
\includegraphics[width=2.5cm]{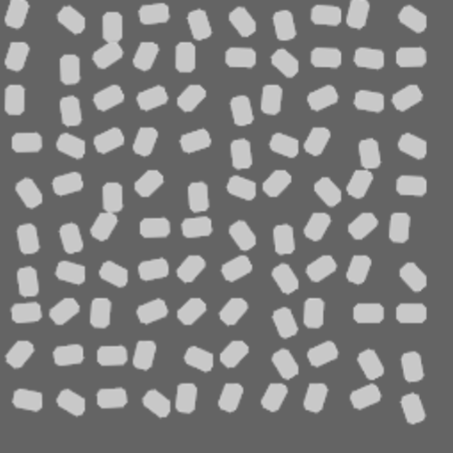}}
\quad
\subfigure[Primal residual convergence for 10\% sampling]{%
\includegraphics[width=4cm]{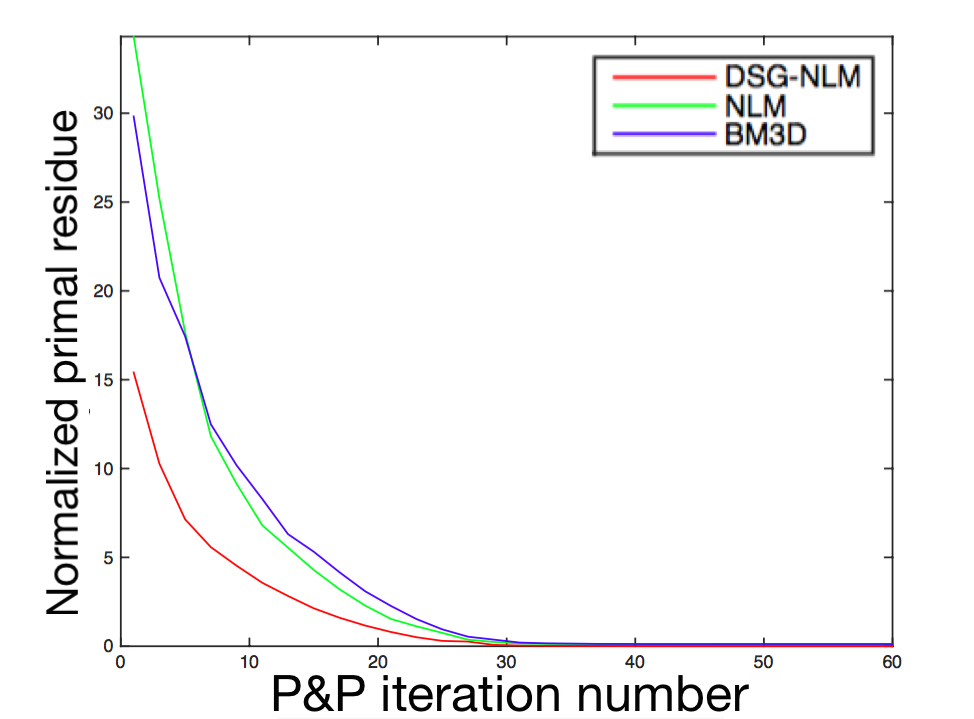}}
\quad
\subfigure[Dual residual convergence for 10\% sampling]{%
\includegraphics[width=4cm]{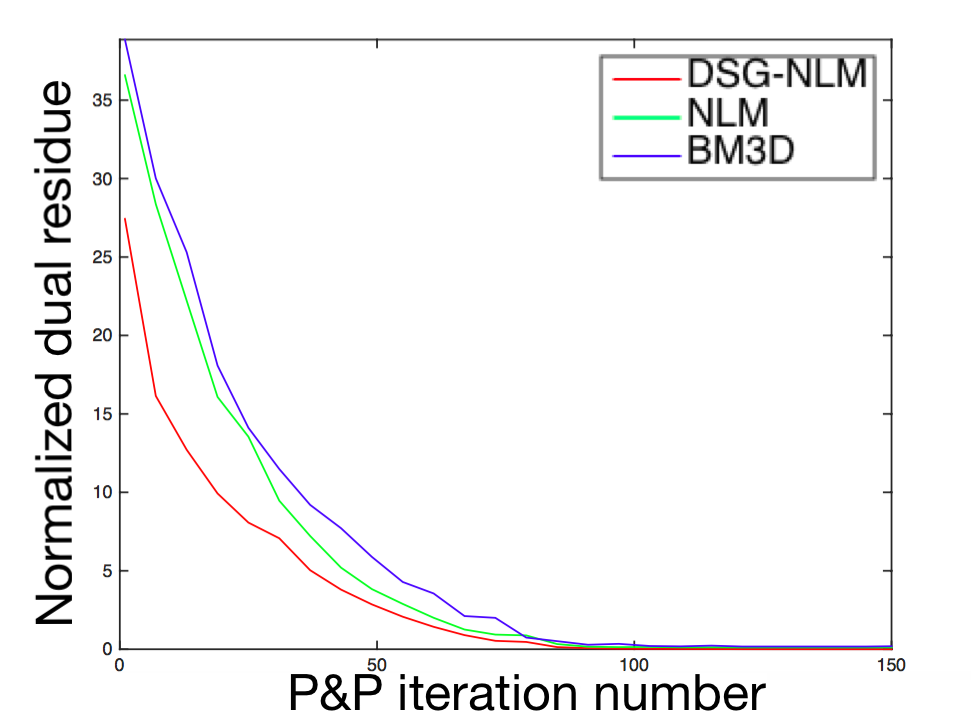}}

\caption{Interpolation of a $256\times 256$ grayscale image of a set of super ellipses.}
\label{fig:Ellipses}
\end{figure*}

\begin{figure*}[!htbp]
\centering

\subfigure[$5\%$ sampling]{%
\includegraphics[width=2.5cm]{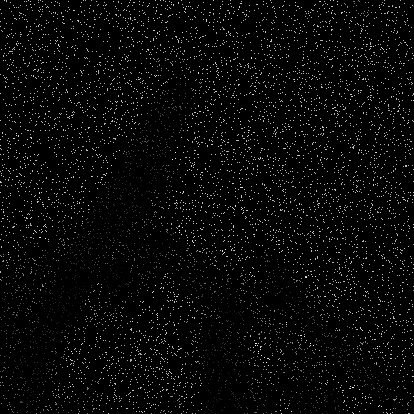}}
\quad
\subfigure[Shepard $5\%$]{%
\includegraphics[width=2.5cm]{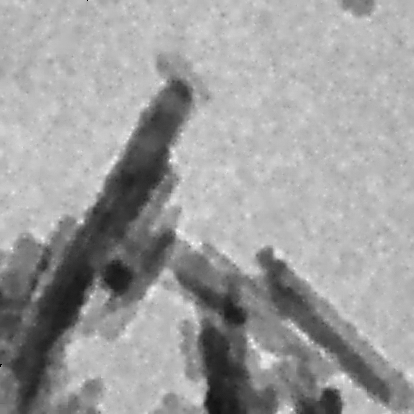}}
\quad
\subfigure[NLM $5\%$]{%
\includegraphics[width=2.5cm]{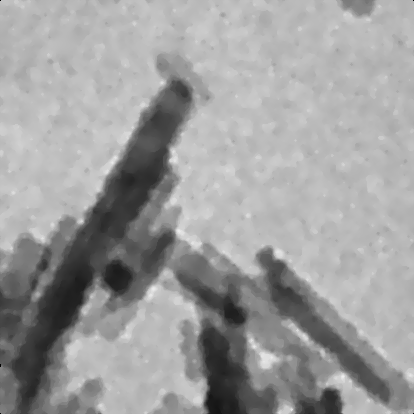}}
\quad
\subfigure[DSG-NLM $5\%$]{%
\includegraphics[width=2.5cm]{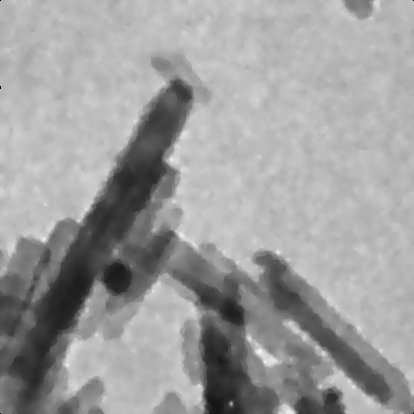}}
\quad
\subfigure[BM3D $5\%$]{%
\includegraphics[width=2.5cm]{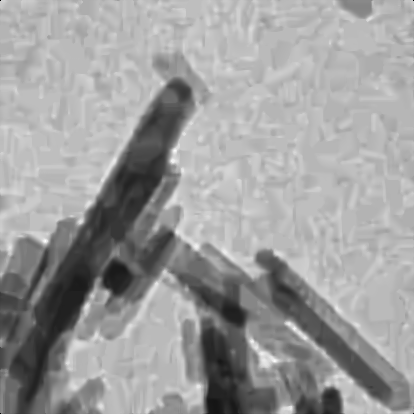}}

\subfigure[$10\%$ sampling]{%
\includegraphics[width=2.5cm]{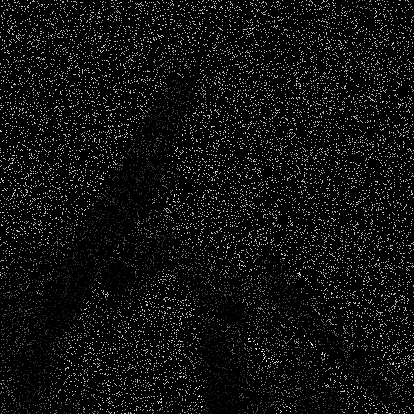}}
\quad
\subfigure[Shepard $10\%$]{%
\includegraphics[width=2.5cm]{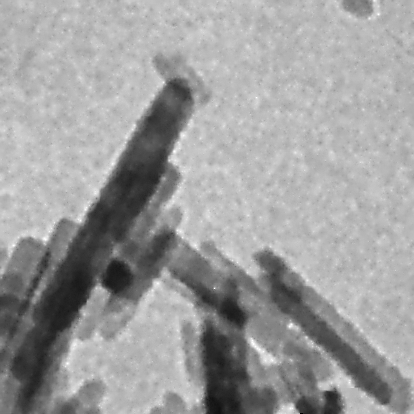}}
\quad
\subfigure[NLM $10\%$]{%
\includegraphics[width=2.5cm]{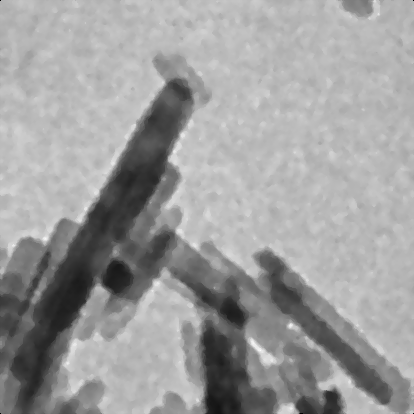}}
\quad
\subfigure[DSG-NLM $10\%$]{%
\includegraphics[width=2.5cm]{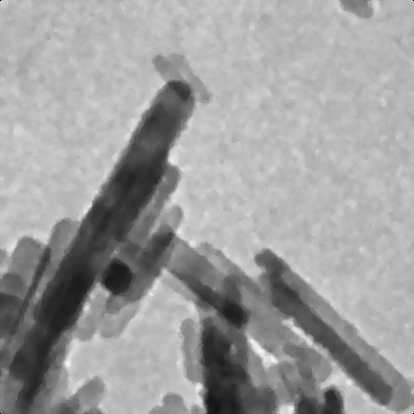}}
\quad
\subfigure[BM3D $10\%$]{%
\includegraphics[width=2.5cm]{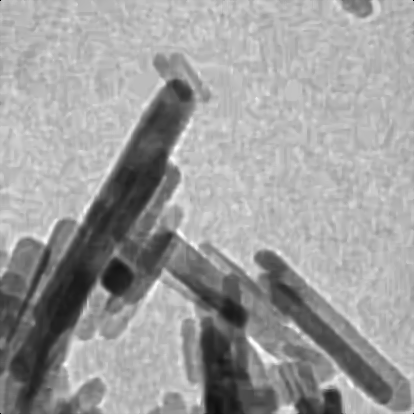}}

\subfigure[Ground truth -- full view]{%
\includegraphics[width=2.5cm]{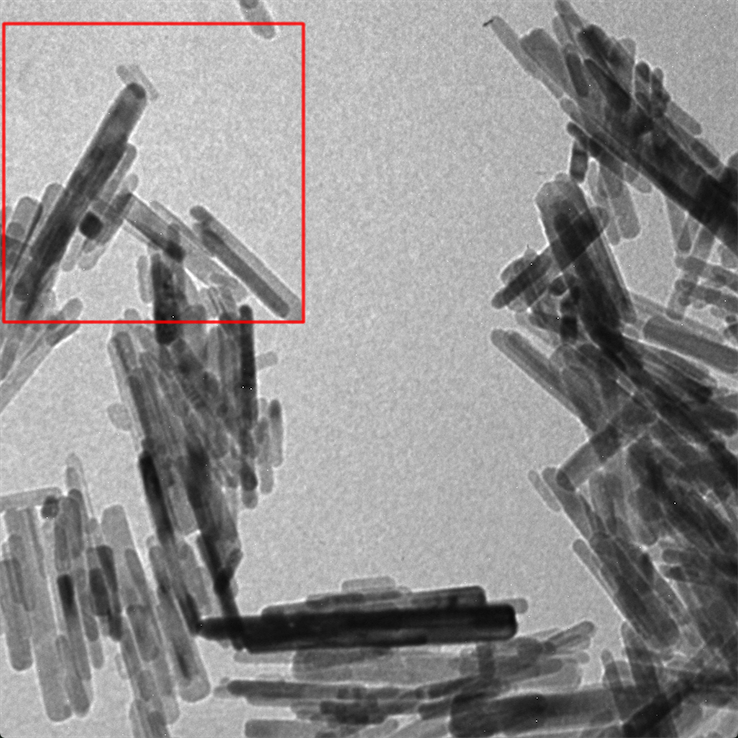}}
\quad
\subfigure[Ground truth -- zoomed into the red box]{%
\includegraphics[width=2.5cm]{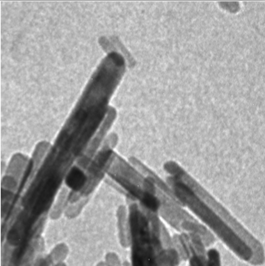}}
\quad
\subfigure[Primal residual convergence for 10\% sampling]{%
\includegraphics[width=4cm]{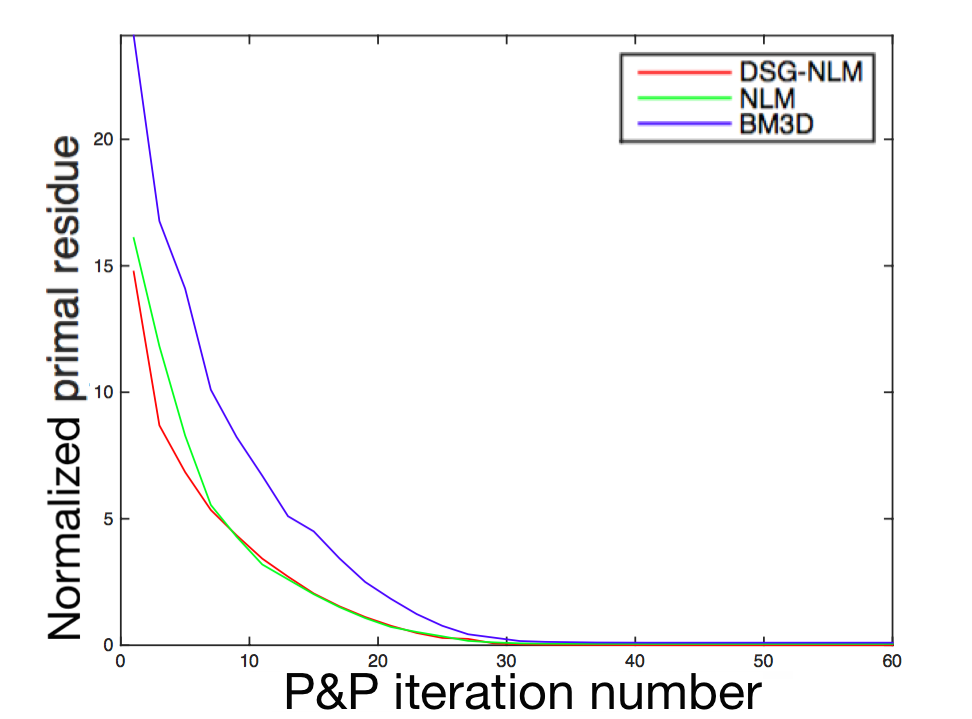}}
\quad
\subfigure[Dual residual convergence for 10\% sampling]{%
\includegraphics[width=4cm]{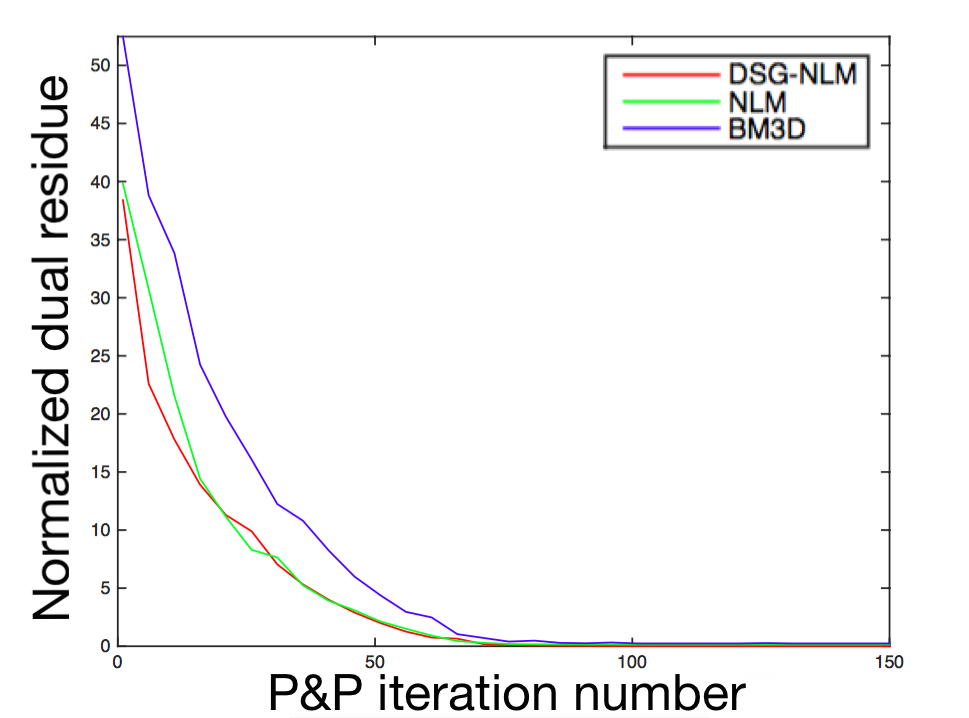}}

\caption{Interpolation of a $414\times 414$ grayscale image of zinc oxide nano-rods.}
\label{fig:zinc_oxide_nanorods}
\end{figure*}


\appendices
\section{Proof of Plug and Play Convergence Theorem}
\label{sec:AppendixPNPProof}

This appendix provides a proof of Theorem~\ref{theorem:PNP-Convergence}.
We start by defining a proximal mapping as any function $H:\mathbb{R}^N \rightarrow \mathbb{R}^N$
which can be expressed in the form
\begin{equation}
H(x) = \arg \min_{v \in \mathbb{R}^N} \left\{ \frac{\| x -v\|^2}{2} + s(v)  \right\} \ ,
\end{equation}
where $s:\mathbb{R}^N \rightarrow \mathbb{R} \cup \{ +\infty \}$ is a proper closed convex function on $\mathbb{R}^N$.
With this definition, we can formally state the theorem proved by Moreau in 1965 \cite{moreau1965proximite}
which gives necessary and sufficient conditions for when $H$ is a proximal mapping.

\begin{theorem}
\label{theorem:1} (Moreau 1965 \cite{moreau1965proximite}) 
A function $H: \mathbb{R}^N \rightarrow \mathbb{R}^N$
is a proximal mapping if and only if\\
(1) $H$ is non-expansive and, \\
(2) $H$ is the sub-gradient of a convex function 
\mbox{$\phi : \mathbb{R}^N \rightarrow \mathbb{R}$}. 
\end{theorem}

In fact, if there exists a function $\phi : \mathbb{R}^N \rightarrow \mathbb{R}$
such that $\forall x \in \mathbb{R}^N$
$$
H(x) = \nabla \phi (x) \ ,
$$
then we say that $H(x)$ is a conservative function or vector field.
The concept of conservative functions is widely used on electromagnetics, for example.
The next well known theorem (see for example \cite[Theorem~2.6, p. 527]{vector_calculus_book}).
gives necessary and sufficient conditions for a continuously differentiable function to be conservative on $\mathbb{R}^N$.

\begin{theorem}
\label{theorem:2} 
Let $H: \mathbb{R}^N \rightarrow \mathbb{R}^N$
be a continuously differentiable function.
Then $H(x)$ is conservative  
if and only if $\forall x\in \mathbb{R}^N$, 
$\nabla H ( x) = \left[  \nabla H ( x) \right]^t$.
\end{theorem}

In general, the sum of two proper closed convex functions, $h = f+g$,
is not necessarily proper. 
This is because the intersection of the two sets 
$A = \{ x \in \mathbb{R}^N: f(x) \leq \infty \}$
and
$B = \{ x \in \mathbb{R}^N: g(x) \leq \infty \}$
might be empty.
Therefore, the following lemma will be needed in order to handle the addition of proper closed convex functions.

\begin{lemma}
\label{theorem:3} 
Let $f$ and $g$ both be proper closed convex functions
and let $h = f +g$ be proper.
Then $h$ is proper, closed, and convex.
\end{lemma}

{\em Proof:}
A proper convex function is closed if and only if it is lower semi-continuous.
So therefore, both $f$ and $g$ must be lower semi-continuous.
This implies that $h$ is also lower semi-continuous.
Since $h$ is formed by the sum of two convex function, it must be convex.
Putting this together, $h$ is proper, convex, and lower-semi-continuous,
and therfore it must be closed.
Therefore, $h$ is a proper, closed, and convex function on $\mathbb{R}^N$.

Using these results,
we next provide a proof of Theorem~\ref{theorem:PNP-Convergence}.

{\em Proof:}
Without loss of generality, we will assume $\beta =1$ and $\sigma_n^2=1$ in 
order to simplify the notation of the proof.

We start by showing result~1 of the theorem,
that $H$ is a proximal mapping
for some proper, closed, and convex function $s(x)$.
To do this, we use Moreau's result stated above in Theorem~\ref{theorem:1}.
In order to meet the conditions of Moreau's theorem,
we first show that $H$ is the sub-gradient of a convex function $\phi: \mathbb{R}^N \rightarrow \mathbb{R}$.
Since $\nabla H(x)$ is assumed to be a doubly stochastic matrix,
we know that $\nabla H ( x) = \left[  \nabla H ( x) \right]^t$.
Then by Theorem~\ref{theorem:2} above, we know that $H(x)$ is conservative
and there must exist a function $\phi$ so that
$$
H(x) = \nabla \phi (x) \ .
$$
Furthermore, since $\nabla H (x) $ is a doubly stochastic matrix,
it must have real eigenvalues in the range $(0,1]$.
Since the eigenvalues are positive, $\phi$ must be convex.
Furthermore, since the eigenvalues are $\leq 1$, $H$ must also be non-expansive.
So therefore, we know that $H$ is a proximal mapping of some proper, closed, and convex function $s(x)$.
More specifically, we know that there exists a proper, closed, and convex function, $s(x)$, on $\mathbb{R}^N$
such that $H$ can be expressed as
\begin{equation}
H(x) = \arg \min_{v \in \mathbb{R}^N} \left\{ \frac{\| x -v\|^2}{2} + s(v)  \right\} \ .
\label{ProximalMapping}
\end{equation}

We next show result~2 of the theorem,
that a MAP estimate exists.
This is equivalent to saying that the function $h(x) = l(x) + s(x)$ takes on its global minimum value
for some $x = \hat{x}_{MAP}$.

First define the function $h(x) = l(x) + s(x)$.
By condition~3 of Theorem~\ref{theorem:PNP-Convergence}
there exists an $x$ and $y$ such that $y = H(x)$ and $l(y) < \infty$.
Since, $y = H(x)$ we also know that $s(y) < \infty$.
Therefore, $h(y) < \infty$ and $h$ is proper.
By Lemma~\ref{theorem:3}, $h$ must also be proper, closed, and convex.

Now to show that $h(x)$ takes on its global minimum,
we need only show that there exists an threshold $\alpha\in \mathbb{R}$
such that the sublevel set of $h$ is a non-empty compact set,
that is 
$$
A_\alpha = \{ x\in \mathbb{R}^N : h(x) \leq \alpha \}
$$
is a non-empty compact subset of $\mathbb{R}^N$.
Since $h$ is a closed function, $A_\alpha$ must be a closed set.
Therefore, it is only necessary to show that $A_\alpha$ is nonempty and bounded.

Define
$$
p^* = \inf_{x\in \mathbb{R}^N} h(x) \ .
$$
Then since $h(x)$ is proper, closed, and convex, 
we know that $\infty > p^* > -\infty$.
Select any $\alpha > p^*$.
So clearly, $A_\alpha$ is nonempty.

Next we show that $A_\alpha$ is bounded.
Since $s(x)$ is a proper closed convex function,
we know that it must have an affine lower bound,
i.e., there exist a finite row vector $b$ and constant $c$ so that
for all $x \in \mathbb{R}^N$
$$
s(x) \geq b x + c \ .
$$
By condition~4 of Theorem~\ref{theorem:PNP-Convergence}, it is always possible to choose $r>1$ so that 
$$
\frac{ f(r) }{ r } > \| b \| + \| c \| + \alpha  \ .
$$
In this case, it is easy to show that for all $\| x\| > r$,
we have that
\begin{eqnarray*}
h(x) 
  &=& l(x) + s(x) \\
  &\geq& f(r) - \left\{  \|b\| r + |c| \right\} \\
  &\geq& r \left\{ \| b \| + \| c \| + \alpha \right\} - \left\{  \|b\| r + |c| \right\} \\
  &\geq& \alpha.
\end{eqnarray*}
So therefore, we know that $\forall x\in A_\alpha$,
$\|x\| < r$, and that $A_\alpha$ is a nonempty bounded and therefore compact set.
Consequently, $h$ must take on its global minimum value
for some value $\hat{x}_{MAP}$ in the compact set $A_\alpha$.

Finally, we show result~3 of the theorem,
that the plug-and-play algorithm convergences.
Since the plug-and-play algorithm is just an application
of the ADMM algorithm, we can use standard ADMM convergence theorems.
We use the standard theorem as stated in \cite[p. 16]{boyd2011distributed}.
This depends on two assumptions.
The first assumption is that $l(x)$ and $s(x)$ must be a proper, closed, and convex functions,
which we have already shown.
The second assumption is that the standard (un-augmented) Lagrangian must have a saddle point.

\noindent The standard Lagrangian for this problem is given by,
\begin{equation}
L(x, v; \lambda) = l(x) + s(v) + \lambda^t(x-v) \ ,
\end{equation}
and the associated dual function is denoted by
$$
g(\lambda ) = \inf_{x,v \in \mathbb{R}^N} L(x, v; \lambda) \ .
$$
We say that $x^* \in \mathbb{R}^N$, $v^* \in \mathbb{R}^N$,
$\lambda^* \in \mathbb{R}^K$ are a saddle point if
$$
L(x, v; \lambda^*) \geq L(x^*, y^*; \lambda^*) \geq L(x^*, y^*; \lambda) \ .
$$
Now we have already proved that a solution to our optimization problem
exists and is given by $x^* =v^* =\hat{x}_{MAP}$.
So we know that the primal problem has a solution given by
\begin{eqnarray*}
p^* 
  &=& \mathop{\inf_{x, v \in \mathbb{R}^N }}_{ v = x } \{ l(x) + s(v) \} \\
  &=& l(x^*) + s(v^* )  \ .
\end{eqnarray*}

Now the pair $( x^*, v^*)$ is a strictly feasible solution to the constrained
optimization problem because  $x^*$ and $v^*$ meet the constraint
and they both fall within the open set $\mathbb{R}^N$.
This means Slater's conditions hold,
and by Slater's theorem, strong duality must also hold for some $\lambda^*$ \cite{Slater1950, BoydVandenberghe2009}.
More specifically, we know that there must exist a $\lambda^*\in \mathbb{R}^N$
such that 
$$
p^* = g( \lambda^* ) \ .
$$

Using this result, we have that
\begin{eqnarray*}
L( x^* , v^* ; \lambda^* ) 
  &=& l(x^*) + s(v^*) + [\lambda^*]^t (x^* - v^* ) \\
  &=& l(x^*) + s(v^*) \\
  &=& p^* = g( \lambda^* ) \\
  &\leq & L( x, v; \lambda^* ) .
\end{eqnarray*}
So we have that $L( x, v; \lambda^* ) \geq L( x^* , v^* ; \lambda^* )$.
Furthermore since $x^* = v^*$, we know that $L(x^*, v^*; \lambda^*) = L(x^*, v^*; \lambda)$
for all $\lambda$.
So putting together these two results,
we have that $L(x, v; \lambda^*) \geq L(x^*, y^*; \lambda^*) \geq L(x^*, y^*; \lambda)$, 
thus proving the existence of a saddle point of the un-augmented Lagrangian, $L(x, v; \lambda)$.

Adapting the theorem of \cite[p. 16]{boyd2011distributed},
we then have the stated convergence results of equation~(\ref{eq:ADMM_convergence}).
\begin{eqnarray}
\lim_{k\to\infty}\{x^{(k)} - v^{(k)}\} = 0;\\
\lim_{k\to\infty} \{l(x^{(k)}) + s(v^{(k)})\} = p^*.
\end{eqnarray}

\section*{Acknowledgment}
The authors thank Gregery Buzzard, professor and head of the Mathematics department 
at Purdue University, for many useful discussions regarding the conditions of convergence of the plug-and-play algorithm. 
They would also like to thank Marc DeGraef, professor of material science at Carnegie Mellon University, for providing simulated aluminum spheres tomography datasets.

\ifCLASSOPTIONcaptionsoff
  \newpage
\fi

\bibliographystyle{IEEEtran}
\bibliography{TCI_2015}

\end{document}